# Towards Human-Like Interaction: A Structured Psychological Computation Framework for Persistent Personality, Dynamic Affect, and Experiential Learning

Haotian Lan

## Abstract

Personality-conditioned large language models can produce coherent social behavior, yet it remains difficult to examine how persistent psychological characteristics contribute to situated decisions and develop through experience. We present the Psychological Modeling Module (PMM), a stateful psychological computing framework operating independently of the general-purpose language model responsible for linguistic interaction. PMM maintains structured personality, affective, cognitive, and experience-related states and connects them to candidate-action evaluation, behavioral self-monitoring, and constrained long-term updating.

Five experiments investigated personality conditioning, cognitive appraisal, affect-related behavior, record-grounded self-evaluation, and longitudinal development. Across 457 complete observations in E01–E04, PMM achieved social-task quality comparable to persona prompting; 62 of 64 matched decisions retained the same core action, while PMM produced final responses that were 67.98% shorter on average, with substantially greater internal token use. Matched personality contrasts revealed trait-specific contributions to action evaluation despite frequent convergence in final choices. The affective experiment identified variation in expression and appraisal alongside discrepancies between assigned and effective runtime states. Record-grounded self-evaluation differentiated maintained choices, implementation adjustments, and changes in core decisions. In 48 longitudinal trajectories comprising 576 events, experience-dependent personality updates preserved initial individual differences and produced additional changes in responsibility planning relative to memory-only learning.

Together, these results establish a computational approach for tracing persistent psychological conditions through decision processes and examining their interaction with short-term dynamics, behavioral monitoring, and experiential change.

## 1. Introduction

### 1.1 Research Background

Recent advances in large language models (LLMs) have enabled artificial agents to generate responses displaying distinct individual characteristics through role descriptions, personality prompting, and conversational histories [1–3]. These capabilities have expanded the possibilities for personalized agents and sustained human–AI interaction, while raising a more fundamental question: can personality expressed in language be developed into a computational mechanism characterized by individual continuity, contextual adaptability, and experience dependence?

Personality psychology provides a theoretical basis for addressing this question. Cognitive–Affective Personality System (CAPS) theory describes individual differences through relatively stable patterns of cognitive, affective, and behavioral responses across situations [4]. Whole Trait Theory further connects enduring traits to fluctuating personality states and the social-cognitive mechanisms that generate them [5]. These perspectives accommodate both continuity in individual characteristics and variability in situated behavior. Enduring dispositions may influence event interpretation, goal evaluation, and action preferences, while current affect, social relationships, and contextual evidence contribute to specific decisions. Repeated experiences may also shape expectations and behavioral tendencies over longer periods, as described in process-oriented accounts of personality development [6].

Research on LLM-based agents has examined personality expression and assessment through trait conditioning, questionnaires, and character interviews [1–3]. Memory-augmented agents have also demonstrated the value of retrieving past experiences, generating reflections, and incorporating feedback into subsequent behavior. Generative Agents, for example, combines memory, reflection, and planning to support extended social simulations [7], while Reflexion uses verbal feedback from previous attempts to inform later decisions [8]. These developments nevertheless leave an important distinction between the appearance of personality in generated language and the computational influence of persistent psychological states on decision formation. Similar actions may arise from different individual conditions and evaluative weights; behavioral changes following experience may likewise reflect updated memories or relationships without implying a change in long-term personality.

This distinction becomes particularly consequential in sustained interaction. An artificial agent with enduring individual characteristics must retain identifiable psychological tendencies while adapting its appraisals to new information, evaluating changes in its own decisions, and incorporating experience into subsequent behavior. Such capabilities involve psychological processes operating on different timescales and cannot be adequately evaluated through stylistic conformity to a character description alone.

Research on artificial personality therefore requires an explicit computational account linking persistent psychological representations, situated cognition, and action selection. Separately observable internal states and behavioral outcomes are essential for examining how individual continuity, short-term adaptation, and experience-dependent development relate to one another.

### 1.2 Research Question and Objectives

The central research question is: How can psychological computation be implemented independently of a general-purpose language model so that persistent individual characteristics participate in cognitive appraisal and behavioral decisions, while exhibiting measurable continuity and change across dynamic situations and accumulated experiences?

Addressing this question requires examining the relationships among psychological representation, its contribution to decision-making, and its development over time. The influence of enduring personality on event interpretation and candidate-action evaluation provides a basis for investigating individual differences in internal cognition. Changes associated with immediate affect and new contextual evidence reveal how psychological continuity can coexist with adaptive decisions. Sustained interaction further introduces the capacity to evaluate previous actions against recorded evidence and to distinguish memory-based adaptation from longer-term changes in personality and subsequent behavior.

The study consequently aims to establish a structured computational account of personality, affect, cognitive appraisal, reference-grounded self-evaluation, and experiential learning. By examining the correspondence between internal psychological variables and observable actions, it seeks to provide testable foundations for individual continuity, contextual adaptability, and experience-dependent development in artificial agents.

### 1.3 Proposed Approach and Main Contributions

To investigate these questions, we propose a Psychological Modeling Module (PMM), an independent framework for psychological computation and cognitive decision augmentation. PMM separates the maintenance and application of psychological states from final language generation. A general-purpose LLM handles natural-language interaction and the linguistic realization of responses, whereas PMM combines internal semantic reasoning, parameterized psychological computation, and cognitive deliberation to produce structured decision conditions grounded in individual characteristics and the current situation.

The framework distinguishes relatively enduring personality traits from transient affect, slower-changing mood, and experience-related states. Personality–situation interactions and state-update mechanisms connect these variables across timescales. Enduring dispositions contribute to contextual appraisal and candidate-action scoring; recorded decisions provide references for subsequent self-evaluation; and accumulated experiences can inform later decisions through update processes specific to each class of psychological state. This organization enables continuity, adaptation, and development to be examined within a common computational account.

Five complementary experiments evaluate the framework at distinct levels. E01 compares structured personality conditioning with conventional personality prompting in social decisions. E02 examines the contributions of personality to contextual appraisal and composite decision scores. E03 investigates relationships among affective conditions, computed psychological states, behavior, and expression. E04 assesses reference-grounded evaluation of predicted and enacted decisions. E05 uses controlled longitudinal simulations to compare long-term personality updating with memory-based learning. Together, the results identify measurable personality-related contributions to decision evaluation, record-grounded recognition of behavioral changes, and incremental behavioral changes associated with long-term personality learning in selected simulated conditions. The affective experiment also highlights the importance of checking whether assigned emotional conditions are preserved in computed states.

The study contributes a computational separation between persistent psychological representation and language generation, enabling individual conditions to be traced through appraisal and decision processes. Its multi-timescale formulation brings enduring personality, dynamic affect, and experience-related change into a common framework while retaining distinct update conditions. The associated experimental approach extends the assessment of artificial personality beyond verbal style to internal decision contributions, self-evaluation against behavioral records, and longitudinal comparisons that distinguish personality learning from memory accumulation. These contributions provide a methodological foundation for investigating artificial agents with persistent individual differences, contextual adaptability, and experience-dependent development.

## 2. Related Work and Theoretical Foundations

### 2.1 Personality Conditioning and Evaluation in Large Language Models

Research on personality in large language models (LLMs) has progressed from role descriptions toward trait-based simulation and assessment. Personality prompting specifies behavioral dispositions and communicative preferences that can shape an agent's responses across tasks. PersonaLLM examined whether LLM-generated text reflects assigned Big Five traits [1], while InCharacter assessed the personality fidelity of role-playing agents through psychological interviews and scales [2]. Serapio-García et al. subsequently developed a psychometric framework for evaluating and shaping perceived personality traits in LLMs [3]. Together, these studies have moved the evaluation of artificial personality beyond informal impressions of character fidelity toward more explicit psychological constructs.

The resulting measures nevertheless address different aspects of personality expression. Questionnaire responses indicate correspondence with trait-related items; character-fidelity measures assess alignment with an intended persona; and situated decisions additionally involve interpreting evidence, balancing goals, and selecting actions. A difference in personality-conditioned outputs does not, by itself, identify the computational contribution of a trait to the underlying decision. This distinction motivates a closer examination of the processes linking relatively stable individual characteristics to cognition and behavior.

### 2.2 Personality Traits, Situational Conditions, and Cognitive Mechanisms

Personality psychology offers a process-oriented account of how enduring individual differences coexist with behavioral variability. Mischel and Shoda's Cognitive–Affective Personality System (CAPS) explains characteristic if–then patterns through interactions among cognitive and affective units, goals, and expectations in specific situations [4]. Whole Trait Theory similarly connects relatively stable trait distributions with momentary personality states and the social-cognitive mechanisms generating them [5]. These approaches suggest that personality-related differences need not manifest as different actions in every situation: social norms and task constraints may lead individuals with different traits to the same broad choice, even when their appraisals and behavioral priorities differ.

The distinction between enduring dispositions and transient responses also has measurement implications. The revised latent state–trait framework of Steyer et al. separates trait-related variation, occasion-specific influences, and measurement processes [9]. Applied to artificial agents, these perspectives support treating stable personality parameters, situation-dependent states, and observable behavior as related but distinct quantities. They also distinguish learning the mechanism that maps traits and situations to responses from updating an individual agent's long-term traits. This separation is essential for interpreting personality-related computation without mistaking temporary state changes for enduring individual development.

## 2.3 Affective Representation, Dynamics, and Regulation

Affective science provides complementary approaches to representing emotion and its influence on behavior. Russell's circumplex model organizes affect along continuous dimensions, particularly valence and arousal [10]. Scherer's Component Process Model emphasizes the coordinated evolution of appraisal and other emotional response components [11]. Together, these perspectives suggest that emotional experience, event interpretation, and observable expression should be distinguished analytically, even when they are dynamically related.

The temporal properties of affect are particularly relevant to agents operating across extended interactions. Kuppens et al. modeled individual differences in affective dynamics, including tendencies toward a baseline and the variability of affective states [12]. Oravecz et al. developed a hierarchical stochastic differential equation approach to modeling continuous affective change [13]. Such work provides a mathematical basis for describing emotion between discrete events, rather than treating affect as a label assigned anew at each conversational turn.

Emotion regulation introduces a further distinction between internal states and outward behavior. Gross's process model differentiates regulatory strategies such as cognitive reappraisal and expressive suppression, which act at different points in the emotion-generative process [14]. Consequently, a change in the emotional tone of an LLM's response is insufficient evidence that its computed internal affect changed in the intended direction. For personality-conditioned agents, a more informative account must examine how enduring dispositions influence event appraisal and affective dynamics, and how the resulting states contribute to action evaluation and expression.

## 2.4 Memory-Augmented Agents, Reflection, and Metacognitive Evaluation

Memory and feedback mechanisms have extended LLM agents beyond isolated interactions. Generative Agents combines a memory stream, reflection, and planning to support behavior informed by prior experiences in simulated social environments [7]. Reflexion uses verbal feedback and stored reflections from earlier attempts to guide subsequent actions without requiring parameter updates to the underlying LLM [8]. Both approaches demonstrate how information accumulated outside the immediate prompt can influence later decisions.

Memory retrieval, reflection, and evaluation of one's own behavior, however, address different cognitive functions. Retrieved memories provide evidence about previous events; reflection may organize that evidence into higher-level interpretations; behavioral self-evaluation requires comparing earlier predictions or intentions with recorded actions and their circumstances. Nelson and Narens's metacognitive framework distinguishes monitoring cognitive processes from controlling subsequent cognition on the basis of that monitoring [15]. For artificial agents, the distinction separates the ability to produce a plausible retrospective account from the ability to identify actual decision changes accurately and use those evaluations to improve future behavior. It also places evidential fidelity, rather than the fluency of reflective language alone, at the center of assessing self-evaluation.

## 2.5 Experiential Learning and Long-Term Personality Development

Experience-dependent behavior can arise through several mechanisms operating at different timescales. Updated memories alter what an agent knows about previous events; repeated encounters may change relationship expectations or habitual responses; and longer-term learning may affect the relatively enduring conditions used to evaluate future situations. In personality psychology, the TESSERA framework links repeated short-term situation–state–reaction sequences with processes of longer-term trait development [6]. It therefore offers a conceptual basis for connecting accumulated experience with enduring individual differences without equating every immediate response with personality change.

For artificial agents, distinguishing these mechanisms is an empirical challenge. Greater caution after repeated unreliable cooperation could reflect new evidence about a particular partner, a revised relationship state, or a more general change in behavioral disposition. Similar observable decisions are compatible with these different explanations. Establishing an incremental role for long-term personality learning consequently requires comparisons in which memory and other experience-related states are accounted for, alongside longitudinal observations of trait representations and later behavior. Such comparisons also need to distinguish short-term affective recovery from slower changes in enduring traits.

### 2.6 Research Positioning

The literature provides substantial foundations for artificial personality, but across different levels of analysis. Personality-conditioned LLM studies characterize the generation and measurement of individual differences [1–3]; personality theories explain why stable dispositions can coexist with context-sensitive behavior [4,5,9]; affective models describe event-related and continuous-time psychological change [10–14]; and memory-based agents demonstrate the influence of accumulated experience and reflective feedback on subsequent action [7,8]. These contributions are complementary, although they do not all address the same computational or evidential question.

The present study connects these lines of work through a psychological computation framework that maintains personality and dynamic affect as distinct internal states, incorporates them into cognitive appraisal and action evaluation, and separates psychological decision formation from final language generation. Its central concern is the relationship between internal psychological contributions, situated behavior, reference-grounded self-evaluation, and experience-dependent individual development. The accompanying experiments examine these relationships at different levels, including a longitudinal comparison designed to distinguish the behavioral contribution of personality updating from memory-based adaptation. In doing so, the study positions persistent artificial personality as a computational and empirically testable account of individual continuity, contextual responsiveness, and change over time.

## 3. Structured Psychological Computation (Methodology)

### 3.1 Theoretical Foundations and Problem Formulation

We propose the Psychological Modeling Module (PMM), a framework for psychological computation and cognitive decision augmentation that makes persistent individual characteristics explicit conditions of situated appraisal and action selection. Its conceptual foundations span personality psychology and computational cognition. Cognitive–Affective Personality System (CAPS) theory accounts for characteristic individual differences through situation-dependent patterns of cognitive–affective activation [4]. Whole Trait Theory relates enduring dispositions to momentary personality states and the mechanisms generating them [5]. Research on cognitive architectures provides complementary accounts of working information, long-term memory, and action selection [19,20]. These theories motivate the psychological relationships to be represented; they do not prescribe the neural functions or update rules specified below. We therefore distinguish theoretical motivation, computational formulation, and empirical validation throughout this chapter.

PMM is architecturally separate from the general-purpose large language model (General LLM). The General LLM handles natural-language interaction and the linguistic realization of responses; PMM maintains psychological states across turns and determines structured decisions from current evidence. Internal semantic analysis, numerical psychological computation, and cognitive deliberation may be implemented using distinct models or algorithms, none of which is identified with the external language generator. Separating decision formation from expression makes it possible to inspect the contributions of personality, affect, relationships, and experience without inferring those contributions solely from response style.

Given an interaction event $x_t$, memory available at the beginning of the turn $M_t^-$, and the prior psychological state $S_t^-$, PMM produces a structured decision $d_t$. The external language generator then realizes this decision using a constrained expression context $\rho_t$:

$$d_t = \mathrm{PMM}_\psi(x_t, M_t^-, S_t^-), \qquad y_t = G_\varphi(x_t, d_t, \rho_t). \quad (1)$$

Equation (1) specifies the division of responsibilities rather than guaranteeing that every General LLM will faithfully express an assigned decision. An implementation must verify the selected action, permitted expression scope, and factual constraints, while logging failures, fallbacks, and actual execution outcomes separately.

## 3.2 Computational Architecture and Evidence-to-Decision Flow

The online PMM pathway combines evidence-grounded event appraisal, psychological-state dynamics, trait-conditioned numerical computation, and decision formation. Memory maintenance and learning follow separate, evidence-gated paths after the interaction. This organization draws on the division of labor among working memory, long-term memory, and control in cognitive architectures [19,20], as well as the reasoning–action organization of language agents [21]. **Figure 1** presents the full architecture and distinguishes online inference, evidence admission, neural-parameter training, and longitudinal personality estimation by their respective inputs and update conditions.

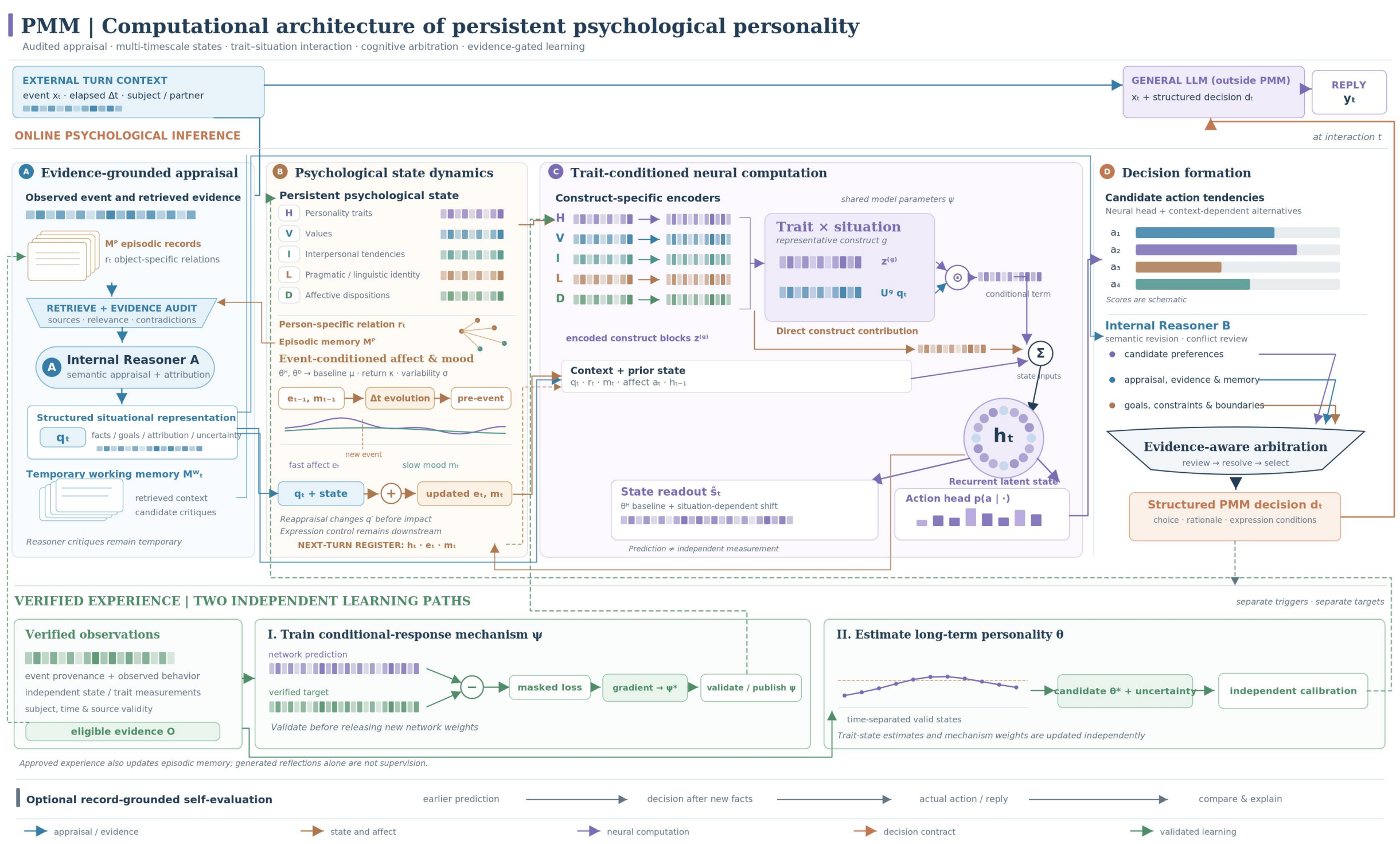

Figure 1. Computational architecture of the Psychological Modeling Module (PMM). The external General LLM is outside PMM. The online pathway comprises (A) evidence-grounded event appraisal, (B) psychological-state dynamics, (C) trait-conditioned numerical computation, and (D) decision formation. Five persistent psychological construct groups are maintained separately from relationship-specific states. The recurrent latent cognitive state feeds the state-readout and action-scoring heads in parallel. Verified experience enters episodic memory and two distinct learning pathways: training the mechanism parameters and estimating longitudinal personality. Record-grounded self-evaluation is an optional post-decision process, not unverified training supervision. Bars, vectors, and trajectories in the figure are schematic representations of computation rather than empirical measurements.

**Event interpretation and evidence audit (A).** Incoming information is represented as a structured event containing an identifier, subject, timestamp, original content, and contextual scope. Relevant episodic memories are retrieved together with their provenance and verification status. Internal Reasoner A identifies known facts, goals, attributions, conflicts, and uncertainties, yielding a structured situational appraisal $q_t$. Unverified interpretations remain in temporary working memory rather than being written directly into persistent memory. Generative Agents and retrieval-augmented generation demonstrate how previous information can inform subsequent reasoning [7,22], but retrieval relevance alone does not establish factual validity. PMM therefore places source verification and evidence admission between retrieval and persistent-state updates.

**Psychological-state dynamics (B).** The current appraisal, preceding affect and mood states, and elapsed real-world time enter the affective dynamics module. Time-dependent propagation precedes the event-induced impulse, producing updated affective states for the current turn. Enduring personality, values, and regulatory dispositions condition the dynamics, while relationship-specific information remains separate from the agent's own persistent traits. This organization is informed by work on emotional appraisal and affect dynamics [11–14].

**Trait-conditioned neural computation (C).** Persistent psychological representations are encoded in five construct groups: traits (H), values (V), interpersonal dispositions (I), linguistic habits (L), and affect-regulatory dispositions (D). Each contributes both directly and through trait–situation interactions to the recurrent cognitive state $h_t$. The state-readout head estimates observable psychological states, whereas the action head directly evaluates candidate behaviors. Both heads share the latent representation, but **state measurement is not a prerequisite for action scoring**. The parallel outputs make state-estimation fidelity and decision contribution separately testable.

**Cognitive deliberation and decision formation (D).** Internal Reasoner B reviews candidate interpretations, anticipated consequences, and conflicting evidence. A Coordinator integrates contextual appraisal, numerical psychological contributions, and review outcomes; a DecisionPort then confirms $d_t$ among candidates satisfying factual and authorization constraints. The General LLM receives the structured decision and only the information authorized for expression. Its generated text and actual delivery outcome are stored separately from the internal decision to allow tests of expression fidelity.

**Evidence admission and learning.** Post-interaction feedback undergoes checks for identity, provenance, timing, and evidential validity before qualifying for episodic memory or supervised training. Temporary working information does not automatically become persistent evidence; self-generated explanations are likewise not independent supervisory ground truth. The distinction between working-memory integration and longer-term learning is informed by the episodic-buffer account [23] and complementary learning systems theory [24]. At the bottom of Figure 1, verified supervision updates the conditional-response network parameters $\psi$, whereas independent longitudinal personality observations support candidate updates and calibration of individual traits $\theta$. These pathways have different targets and admissibility conditions.

To make these flows auditable, computation records should preserve event identity and evidence provenance, state snapshots and versions, candidate actions and decomposed scores, the final structured decision, expressed language and execution receipt, and the scope of any subsequent update. These data contracts define the proposed methodology; whether every contract is implemented in a particular software release must be established through version-specific logs and tests.

### 3.3 Multi-Timescale States and Personality–Situation Interaction

The model indexes individuals by $i$, interaction turns by $t$, and slower longitudinal phases by $k$. Its persistent psychological state includes enduring personality, transient affect, slower mood, relationship state, situational appraisal, recurrent cognition, and episodic memory:

$$S_{i,t} = \{\theta_{i,k}, e_{i,t}, m_{i,t}, r_{i,t}, q_{i,t}, h_{i,t}, M_{i,t}\}. \quad (2)$$

Personality is encoded in the H/V/I/L/D construct groups. Missing or unmeasured dimensions retain validity masks and uncertainty rather than being treated as reliable numerical assessments merely because a language model can describe them. Individual parameters $\theta$ and the shared conditional-response parameters $\psi$ are maintained separately: the former represent a specific individual, while the latter govern the general mapping from psychological conditions and situations to responses.

CAPS emphasizes the conditional nature of personality-related responses [4], while Whole Trait Theory distinguishes enduring traits from dynamic states [5]. Motivated by these accounts, PMM generates the latent cognitive state through a bilinear interaction between psychological constructs and current context:

$$h_t = \tanh(Ah_{t-1} + Bq_t + Rr_t + W_m m_t + W_e e_t + b + \Psi_t).$$

$$\Psi_t = \sum_{g\in\{H,V,I,L,D\}} \left[W_g z_t^{(g)} + C_g \left(z_t^{(g)} \odot U_g q_t\right)\right]. \quad (3)$$

Here $z_t^{(g)}$ denotes the encoded construct group, and $A$, $B$, $R$, $W_m$, $W_e$, and $W_g$, $C_g$, $U_g$ are corresponding model weights; $\odot$ denotes elementwise multiplication. The preceding cognitive state, current appraisal, relationship information, and affective conditions enter through distinct terms. The interaction allows a given disposition to contribute differently across situations. **The nonlinear mapping and network architecture are proposed computational choices, not equations derived directly from CAPS or Whole Trait Theory.**

A state-readout head and an action head produce separate outputs from the shared $h_t$. Informed by the revised state–trait framework [9], the relationship between momentary personality state and its external observation is expressed as:

$$s_t = \Lambda\theta_k^{(H)} + H_s h_t + u_t, \qquad y_t^{\text{obs}} = L s_t + \varepsilon_t. \quad (4)$$

Here $s_t$ is the current personality state, $u_t$ represents other occasion-specific variation, $y_t^{\text{obs}}$ denotes a behavioral or psychometric observation, and $\varepsilon_t$ is measurement error. The observation $y_t^{\text{obs}}$ is distinct from the linguistic output $y_t$ in Equation (1). Longitudinal comparisons further require measurement comparability for the same construct across occasions [28]. Accordingly, latent-state changes, observed behavioral changes, and revisions to enduring traits require distinct evidence.

## 3.4 Affective Dynamics, Event Impulses, and Regulation

Affective computation must account for continuous evolution between interactions and changes triggered by meaningful events. Research on individual differences in affective baselines and recovery [12] and continuous-time affective modeling [13] supports the temporal component. Scherer's component-process account [11] and Smith and Ellsworth's appraisal research [27] motivate the separate event-related component. PMM therefore models elapsed-time propagation, appraisal-induced impulses, and expressive regulation as distinct processes.

Conditional on enduring traits and affect-regulatory dispositions, the individual's affective dynamics parameters are defined as:

$$\mu_i = g_\mu\big(\theta_i^{(H)}, \theta_i^{(D)}\big),$$

$$\kappa_i = \text{softplus}\Big(g_\kappa\big(\theta_i^{(H)}, \theta_i^{(D)}\big)\Big) + \epsilon_\kappa,$$

$$\sigma_i = \text{softplus}\Big(g_\sigma\big(\theta_i^{(H)}, \theta_i^{(D)}\big)\Big). \quad (5)$$

Here $\mu_i$ is the latent affective baseline, $\kappa_i$ is the mean-reversion rate, and $\sigma_i$ controls stochastic diffusion. Using the exact discrete-time transition of an Ornstein–Uhlenbeck (OU) process [16], the state after an elapsed real-time interval $\Delta t$ between events is:

$$x_{t+\Delta t} = \mu_i + \exp(-\kappa_i \Delta t)(x_t - \mu_i) + \varepsilon_{\Delta t},$$

$$\varepsilon_{\Delta t} \sim \mathcal{N}\left(0, \frac{\sigma_i^2}{2\kappa_i}\left[1 - \exp(-2\kappa_i \Delta t)\right]\right). \quad (6)$$

Equation (6) adopts the OU transition mathematically; the mappings from personality to $\mu_i$, $\kappa_i$, and $\sigma_i$ are proposed extensions requiring calibration with longitudinal observations. $x_t$ denotes affect on a latent computational scale, which can be mapped into a bounded affective representation by an appropriate link function. Propagation over elapsed time does not imply that any new external event has occurred.

Upon a new event, Reasoner A produces an appraisal of goal relevance, responsibility, controllability, normative conflict, and relational meaning. A parameterized function then produces an emotion-related impulse:

$$j_t = f_\phi\big(q_t, \theta_i^{(H)}, \theta_i^{(V)}, \theta_i^{(D)}, r_t, m_t, e_t^-\big), \quad (7)$$

$$e_t^{\text{post}} = \ell^{-1}\big(\ell(e_t^{\text{pre}}) + j_t\big). \quad (8)$$

Equation (7) uses variables motivated by appraisal research [11,27], while Equation (8) specifies a bounded numerical update. $e_t^{\text{pre}}$ and $e_t^{\text{post}}$ denote affective states before and after the event, and $\ell$ maps between bounded and latent scales. Transient affect and slower mood have different temporal characteristics; mood may integrate influences across several events, but neither state is itself validated evidence of an enduring personality change. Following Gross's emotion-regulation account [14], cognitive reappraisal and expressive suppression are assigned to different processing stages so that internal affect, evaluative adjustment, and outward expression can be assessed separately.

### 3.5 Cognitive Deliberation and Behavioral Decision Computation

The trait-conditioned network evaluates candidate actions from the latent state $h_t$; it does not require converting the state readout into a fixed personality label before making a decision. Action selection combines facts and task goals with personality-related preferences, affect, relationship history, and anticipated outcomes. Multiattribute utility theory provides a formal account of evaluating trade-offs among objectives [29], while adaptive mixtures of local experts demonstrate how input-dependent gating can distribute contributions across computational components [25]. PMM draws on these mathematical principles to combine psychological sources through dynamic decision weights:

$$Q_t(c) = u_{\text{ctx}}(c|q_t) + \sum_{j\in\mathcal{J}} \beta_{t,j}\, u_j(c|h_t, S_t), \quad (9)$$

$$\boldsymbol{\beta}_t = \text{softmax}\left(g_{\psi_\beta}(h_t, q_t)\right). \quad (10)$$

For candidate action $c$, $u_{\text{ctx}}$ summarizes contextual facts, goals, and consequences, while $u_j$ represents the contribution of psychological source $j$; $\beta_{t,j}$ is its weight at the current turn. Training updates the parameters $\psi_\beta$ that generate these weights; the realized $\beta_t$ is a context-dependent output, not a persistent personality parameter. The weighted score is an implementation choice and should not be interpreted as the direct addition of psychometrically calibrated trait scores.

After review by Reasoner B and the Coordinator, the DecisionPort selects an action from the factually and operationally admissible set $\mathcal{C}_t^{\text{valid}}$:

$$d_t = \underset{c\in\mathcal{C}_t^{\text{valid}}}{\text{argmax}}\, Q_t(c). \quad (11)$$

Valid choices can include clarification, postponement, or refusal. A change in a psychological contribution need not change the selected core action: strong constraints or a substantial gap between candidate scores can yield identical choices under different psychological conditions. This is why effects on internal appraisal and effects on observable behavior must be analyzed separately.

Behavioral self-evaluation takes place after the primary decision pathway. It compares earlier predictions, decisions made after additional evidence, and records of actual behavior, assessing both the detected change and its evidential consistency. This function corresponds principally to metacognitive monitoring in the account of Nelson and Narens [15]. E04 tests self-evaluation with explicit historical references; it does not establish unaided introspection or a complete autonomous feedback-control loop.

### 3.6 Episodic Memory and Independent Training of Response Mechanisms

Long-term psychological continuity requires reliable historical evidence, but storing memories, training response mechanisms, and revising an individual's enduring traits are different operations. The episodic buffer and complementary learning systems provide accounts of short-term information integration and longer-term learning [23,24]. Studies of memory and reflection in language agents further show how previous experience can affect later actions without updates to the underlying language model [7,8].

An event record $o_t$ qualifies for the admissible evidence set $\mathcal{O}$ only after checks of subject scope, provenance, time, duplicate identity, and factual status. Persistent-memory admission is separate from parameter training. A generated explanation may remain a provisional interpretation, but it is not an independently verified label for personality or psychological state. During supervised training, model predictions are compared with validated targets, using masks for missing or inadmissible observations.

Drawing on multitask learning with shared representations [26] and neural-network probability calibration [17], the mechanism-training objective combines state prediction, action prediction, calibration, and regularization:

$$\mathcal{L}_{\text{mechanism}} = \lambda_s \mathcal{L}_{\text{state}}^{\text{masked}} + \lambda_a \mathcal{L}_{\text{action}}^{\text{masked}} + \lambda_c \mathcal{L}_{\text{calibration}} + \lambda_r \mathcal{R}(\psi),$$

$$\psi \leftarrow \psi - \eta \nabla_\psi \mathcal{L}_{\text{mechanism}}. \quad (12)$$

Here $\psi$ includes trainable parameters of the construct encoders, conditional-response mechanism, and dynamic decision-weight network; $\lambda$ specifies the relative loss weights and $\eta$ the learning rate. In Figure 1, this pathway proceeds through verified targets, masked losses, candidate mechanism parameters, and validation before deployment. **The resulting update changes the model that generates conditional responses, not the long-term personality values of a particular agent.**

### 3.7 Longitudinal Personality Estimation and Evidence Calibration

The TESSERA framework connects repeated situation–state–reaction sequences with processes of long-term personality development [6], but does not specify universal numerical learning rates that can simply be assigned to an artificial agent. PMM therefore models individual trait learning as a slow state-estimation problem, separate from network training, and aggregates deviations across events supported by valid evidence:

$$\zeta_t = \hat{s}_t - \Lambda\theta_k, \quad (13)$$

$$\bar{\zeta}_k = \frac{\sum_{t\in k} w_t \zeta_t}{\sum_{t\in k} w_t}, \qquad \theta_{k+1}^- = F\theta_k + G\bar{\zeta}_k + b_\theta. \quad (14)$$

Here $w_t$ reflects the quality and independence of an observation, $F$ captures persistence in existing traits, and $G$ maps aggregated deviations into a candidate longitudinal change. Equations (13)–(14) are computational hypotheses motivated by personality-development theory. Short-term deviations or changes in relationships need not reflect enduring personality change; candidate updates require corroboration across time.

When independent personality observations $y_{k+1}^\theta$ are available and a linear-Gaussian approximation is appropriate, Kalman state estimation [18] provides a basis for uncertainty propagation and posterior revision:

$$P_{k+1}^- = FP_k^{\text{post}}F^\top + G\Omega_k G^\top + Q_\theta, \quad (15)$$

$$K_{k+1} = P_{k+1}^- J^\top \left(JP_{k+1}^- J^\top + R_y\right)^{-1}, \quad (16)$$

$$\theta_{k+1}^{\text{post}} = \theta_{k+1}^- + K_{k+1}\left(y_{k+1}^\theta - J\theta_{k+1}^-\right). \quad (17)$$

Here $P_k^{\text{post}}$ is the previous posterior covariance, $\Omega_k$ the uncertainty of the aggregated deviation, $Q_\theta$ the longitudinal process noise, $J$ the observation matrix, and $R_y$ the measurement-error covariance. These equations adapt general state-space estimation; interpreting them as personality measurement additionally requires independent measurement validity and longitudinal invariance [28]. When these conditions are unmet, a personality-change hypothesis should remain provisional rather than being promoted on the basis of an unverified neural output or a single conversational impression.

The second learning pathway in Figure 1 thus forms candidate enduring-trait updates and uncertainty estimates from independent longitudinal evidence, then returns calibrated changes to the persistent construct bank only after validation. Its timescale, supervision, and deployment conditions differ from the mechanism-training pathway in Equation (12).

### 3.8 Correspondence Between the Method and Experimental Evidence

The framework links psychological constructs, numerical operations, and an executable architecture, but defining the full mathematical model does not mean that every component has been validated experimentally. We therefore distinguish whether an internal state was correctly retrieved, whether a psychological variable entered candidate scoring, whether observable behavior changed, and whether accumulated experiences produced sustained changes over time.

E01 examines behavioral performance under structured personality conditioning. E02 evaluates personality-related contributions to internal scores and situational updates, with some runs relying on a deterministic fallback computation. E03 inspects emotion-condition labels, effective runtime states, and behavioral expression separately, avoiding the assumption that accepted input labels constitute successful affective interventions. E04 tests the recognition of behavioral changes against explicit decision records. E05 compares longitudinal behavior in controlled settings with joint personality-and-memory learning against memory-only learning. Its bounded slow trait-update mechanism is not an empirical validation of the full posterior estimator in Equations (15)–(17). The trainable mechanisms and posterior personality calibration require additional independent longitudinal evidence.

Accordingly, the architecture diagram describes computational responsibilities, the equations specify proposed mechanisms, and the experiments provide evidence about the particular pathways that were actually executed. Their correspondence is established through data interfaces, recorded execution paths, and provenance, without equating computational feasibility with psychometric validity.

# 4. Experimental Design and Evaluation

## 4.1 Experimental Framework and Research Questions

Five complementary experiments (E01–E05) were conducted to investigate how structured psychological computation contributes to cognition and behavior in artificial agents. They address personality conditioning, personality-dependent situational judgment, dynamic affect, behavioral self-evaluation, and learning from accumulated experience. The experimental framework draws on theories of personality–situation interaction, affective dynamics, and metacognition [4–6,12–15], with distinct comparisons and evaluation criteria for each computational mechanism.

The analysis integrates observable behavior with internal computational traces. Behavioral evaluation examines contextual understanding, appropriateness of action, and individual differences. Process-level evaluation tracks psychological states, situational appraisals, and contributions to candidate actions. The longitudinal experiment additionally assesses trait parameters, relationship states, and behavioral change across observation periods. Accordingly, the five experiments use paired responses, staged trajectories, matched-condition sets, or longitudinal trajectories as their respective units of analysis.

**Table 1. Overview of experimental designs and evaluation targets**

| Experiment | Research focus | Comparison conditions | Primary evaluation targets |
|---|---|---|---|
| E01 | Personality implementation | Baseline, Persona Prompting, PMM | Task quality, core actions, decision parameters, personality-related differences |
| E02 | Personality in cognitive decision-making | Four A/C profiles; three-stage scenarios | Internal appraisals, trait contributions, action selection |
| E03 | Dynamic affect | Positive, neutral, and negative conditions | Effective affective state, appraisal, expression, and strategy |
| E04 | Behavioral self-evaluation | Initial prediction, actual action, retrospective evaluation | Change detection, causal explanation, evidential fidelity |
| E05 | Long-term experiential learning | Four learning configurations; three experience environments | Stable traits, state trajectories, and longitudinal behavior |

## 4.2 E01: Comparison of Personality Implementation Methods

E01 examines the behavioral expression of personality in complex social decisions and compares PMM with conventional persona prompting. DeepSeek was used as the language model to generate 144 responses across 16 scenarios covering team responsibility, resource allocation, risk management, and social negotiation.

Three conditions were evaluated: a baseline without explicit personality conditioning, Persona Prompting, and PMM. The latter two conditions used four profiles formed by high and low levels of Agreeableness (A) and Conscientiousness (C). Each profile was evaluated on the same 16 scenarios, yielding 64 paired PMM–prompting comparisons. An additional 16 baseline responses served as behavioral references.

Evaluation comprised task quality and personality-related behavior. Task quality was rated on a 0–9 scale covering factual understanding, responsible handling of the situation, and action appropriateness. Behavioral coding distinguished the core action, implementation details, and scenario-specific indicators associated with A and C. Matched comparisons examined differences in action selection, resource allocation, timing, and risk trade-offs, consistent with the conditional behavioral patterns described in personality theory [4,5].

## 4.3 E02: Personality Contributions to Cognitive Appraisal and Decision-Making

E02 investigates how enduring personality characteristics enter situational interpretation, internal appraisal, and action selection. The design comprised nine interpersonal scenarios, four A/C profiles, and three repetitions, producing 108 planned records, of which 100 contained complete three-stage trajectories.

Each trajectory progressed from initial situation interpretation to appraisal following additional information and, finally, an action decision. Threat appraisal, need for clarification, supportive orientation, and personality-related contributions to candidate actions were recorded at each stage to characterize how internal judgments changed as new evidence became available.

Personality contributions were compared within four-profile blocks matched by scenario and repetition. Analyses estimated high–low contrasts for the target A and C dimensions, cross-dimensional contributions, and changes across stages. The corresponding behavioral analysis coded actual interpersonal actions—such as support, clarification, refusal, and boundary maintenance—from the final responses and examined their relationship to internal candidate scores.

Execution traces also recorded the status of the internal reasoning components and the RegistryPMM local deterministic fallback path, allowing action-score contributions to be associated with the computational route that produced them.

### 4.4 E03: Affective Conditions and Behavioral Responses

E03 examines variation in situational appraisal, behavioral strategy, and linguistic expression under positive, neutral, and negative affective conditions. Personality profile, scenario, and repetition were matched across conditions. The dataset comprised 144 original records, including 130 complete staged trajectories and 35 fully matched affective triplets.

Behavioral analysis distinguished four levels: core action, event appraisal, interpersonal boundary, and linguistic expression. This approach follows componential and regulatory perspectives on emotion [11,14] and characterizes where affect-related variation appears in the cognitive–behavioral process.

Internal-state evaluation recorded whether affect injection was accepted, the effective affective vector during execution, affective contributions to candidate actions, and the planned expression strategy. Differences between target and effective states were assessed using state deviations, root-mean-square error (RMSE), and the ordering of effective values across experimental conditions.

Matched comparisons were conducted within the same personality and scenario configurations. For binary expression-strategy variables, Cochran's Q test was used to assess differences among the three related conditions [31]. Internal-state traces were examined alongside behavioral coding to characterize affective processing from state assignment through decision evaluation and final expression.

### 4.5 E04: Prediction, Actual Action, and Self-Evaluation

E04 evaluates an agent's ability to identify and explain changes in its own behavior. Guided by the distinction between metacognitive monitoring and control [15], the experiment compares initial predictions, subsequent decisions, and retrospective evaluations against recorded behavior.

The design included nine interpersonal scenarios, four personality profiles, and three repetitions, producing 108 original records. Of these, 83 contained complete prediction, action, and self-evaluation branches.

In the prediction branch, the agent received the initial scenario and stated its anticipated action. The action branch introduced additional contextual information in stages and recorded the resulting decision. The retrospective branch provided the earlier prediction and actual action record and required the agent to identify behavioral differences and explain their relationship to the new information.

The 83 complete sequences were scored using the archived context-anchored rubric: prediction alignment (0–2), discrepancy recognition (0–3), source fidelity (0–3), and reason explanation (0–3), yielding a maximum total of 11. Prediction–action relationships were also classified as maintained core choice, adjusted implementation, or changed core choice. These complete-sample ratings and classifications were analyzed descriptively and kept distinct from the separately adjudicated semantic reference cases.

Independent manual semantic adjudication was conducted for the ten complete L08 invitation sequences by comparing each prediction, final action, and retrospective self-evaluation. Two further sequences, E04_P3_L02_R1 and E04_P3_L04_R3, were examined as specific cases of factual-reconstruction errors. For all 83 sequences, supplementary lexical analysis calculated cosine similarity between the prediction and the actual Stage 1 and Stage 3 responses in a shared character 2–4-gram TF–IDF feature space. Changes from Stage 1 to Stage 3 were summarized with trial-level bootstrap confidence intervals. The lexical analysis measures differences in wording; semantic judgments of action-change recognition are drawn from the independently adjudicated records.

### 4.6 E05: Long-Term Experiential Learning and Personality Development

E05 investigates how accumulated experience influences enduring personality characteristics, relationship states, and behavioral tendencies, with particular attention to the contribution of trait updating beyond memory-based learning. Drawing on the TESSERA account of repeated experiences and personality development [6], the study constructed longitudinal simulations with controlled environments and learning configurations.

Four initial personality profiles (P1–P4), three experience environments, and four learning conditions yielded 48 longitudinal trajectories. Each trajectory comprised 12 events organized into four observation periods (W1–W4) over 36 simulated days, totaling 576 events.

The environments represented positive cooperation (E1), betrayal and accountability (E2), and short-term affective disruption (E3). The learning conditions were C1, joint updating of enduring personality, memory, and cognitive states; C2, memory and cognitive-state updating with enduring traits frozen; C3, static personality; and C4, persona prompting combined with experiential memory.

Comparisons were matched by initial profile and experience environment. Longitudinal measures included A/C trait parameters, interpersonal trust and cognitive states, short-term affective responses, and behavioral ratings of interpersonal consideration and responsibility-oriented planning on 1–5 scales.

Analyses examined changes from W1 to W4 and contrasts among learning conditions under equivalent experience sequences. The C1–C2 comparison assessed the incremental contribution of enduring-trait updating; C3 provided a static reference, while C4 represented an alternative combination of persona prompting and memory. Psychological states and behavioral outputs were retained at each observation period for longitudinal analysis.

### 4.7 Evaluation Criteria and Statistical Analysis

Each experiment employed measures aligned with its research target. E01 assessed outward personality-related behavior; E02 quantified internal cognitive contributions; E03 examined affective states and behavioral responses; E04 assessed self-monitoring and explanatory fidelity; and E05 evaluated personality development and experience-dependent behavioral change.

Paired and matched analyses preserved scenario, personality, or repetition identifiers. Continuous measures were summarized using means, contrasts, and directional consistency; categorical outcomes were reported as counts, proportions, and condition-specific distributions. Bootstrap confidence intervals were estimated by resampling the relevant matched blocks or longitudinal trajectories [30].

Longitudinal analyses used complete simulated trajectories as observational units, tracking psychological parameters and behavioral ratings across periods and comparing learning conditions and experience environments. Behavioral annotations were anchored to scenario facts and separately characterized core actions, implementation choices, affective expression, and explanatory quality.

The archived dataset contains original interactions, psychological states, candidate-action scores, execution records, and longitudinal learning traces, together with the sources of semantic annotations and model-based evaluations. These records support joint analysis of the internal psychological computations and the observable decisions reported in the Results section.

## 5. Results and Analysis

### 5.1 Dataset Overview

E01–E04 yielded 504 experimental records, of which 457 constituted complete analytical sequences: 144 in E01, 100 in E02, 130 in E03, and 83 in E04. E05 comprised 48 longitudinal experience trajectories with 12 events per trajectory, totaling 576 events. Following the analytical structure of the five original experimental manuscripts, results are reported through task ratings, matched contrasts, internal computational traces, behavioral annotations, and the corresponding result figures.

### 5.2 E01: Decision Performance and Expressive Efficiency under Structured Personality Conditioning

E01 compared an unconditioned baseline, conventional natural-language Persona Prompting, and PMM conditioning across 16 social decision-making scenarios. The dataset contained 16 baseline responses, 64 Persona Prompting responses, and 64 PMM responses. The latter two conditions were paired by personality configuration and scenario.

#### 5.2.1 Overall Performance across Personality Conditioning Methods

**Table 2. Overall performance across personality conditioning methods**

| Evaluation metric | Baseline | Persona Prompting | PMM |
|---|---|---|---|
| Sample size | 16 | 64 | 64 |
| Mean task quality (0–9) | 8.875 | 9.000 | 8.953 |
| Standard deviation | 0.342 | 0.000 | 0.213 |
| Full-score rate (%) | 87.50 | 100.00 | 95.31 |
| Explicit personality description | No | Yes | No |
| Mean system-entry length (characters) | 84 | 381.75 | 103 |
| Mean final response length (characters) | 2,188.69 | 2,502.88 | 801.52 |
| Median final response length (characters) | 2,156 | 2,402 | 769.5 |
| Mean cumulative input tokens | 2,721.50 | 3,332.84 | 94,381.89 |
| Mean cumulative output tokens | 2,558.25 | 2,826.31 | 16,039.66 |

*Note: Final response length is measured in characters in the third-stage user-facing response. Token counts include all model calls within each trial. System-entry length excludes the conditioning context generated automatically by PMM.*

The three methods achieved mean task-quality scores of 8.875, 9.000, and 8.953, respectively, on a nine-point scale. PMM achieved a full-score rate of 95.31%, closely approaching the task quality of Persona Prompting. The more pronounced differences concerned computational resource allocation and response organization. PMM produced a mean final response of 801.52 characters, compared with 2,502.88 characters for Persona Prompting, a reduction of 67.98%. Its mean system-entry text decreased from 381.75 to 103 characters, a 73.02% reduction.

More concise user-facing text was accompanied by greater internal model usage. PMM consumed an average of 94,381.89 cumulative input tokens and 16,039.66 output tokens per trial, compared with 3,332.84 and 2,826.31 under Persona Prompting. Thus, structured psychological conditioning allocated substantially more computation to internal appraisal and decision preparation while generating shorter final responses. Figure 2(a) displays response length and task quality for 64 matched pairs; Figure 2(c) relates cumulative token use to final response length.

#### 5.2.2 Cross-Scenario Task Quality and Matched Comparisons

**Table 3. Decision performance across four social scenario categories**

| Scenario category | Baseline quality | Prompt quality | PMM quality | Core-action agreement (%) | Response length reduction (%) |
|---|---|---|---|---|---|
| Interpersonal cooperation | 9.000 | 9.000 | 8.875 | 87.50 | 65.0 |
| Interpersonal conflict | 8.500 | 9.000 | 9.000 | 100.00 | 70.8 |
| Responsibility and rules | 9.000 | 9.000 | 9.000 | 100.00 | 67.0 |
| Resources and interests | 9.000 | 9.000 | 8.938 | 100.00 | 69.2 |
| Overall | 8.875 | 9.000 | 8.953 | 96.88 | 68.0 |

*Note: Core-action agreement and response-length reduction are calculated for PMM relative to Persona Prompting.*

Task quality remained high across all four scenario categories. Core-action agreement reached 100% for interpersonal conflict and responsibility-and-rules tasks and 87.50% for interpersonal cooperation. Across categories, mean PMM responses were 65.0%–70.8% shorter than those under Persona Prompting. In settings with explicit normative constraints, both approaches commonly selected the same core action, although implementation parameters could differ.

**Table 4. Matched comparisons between PMM and Persona Prompting (64 pairs)**

| Evaluation metric | Matched pairs | Proportion (%) |
|---|---|---|
| Identical task quality | 61/64 | 95.31 |
| PMM quality lower by one point | 3/64 | 4.69 |
| Identical core action | 62/64 | 96.88 |
| Different core actions | 2/64 | 3.13 |
| Shorter system-entry text under PMM | 64/64 | 100.00 |
| Shorter final response under PMM | 64/64 | 100.00 |
| Identical quality with shorter response | 61/64 | 95.31 |
| Identical quality and core action with shorter response | 60/64 | 93.75 |

Of the 64 matched pairs, 61 (95.31%) received identical task-quality ratings and 62 (96.88%) selected the same core action. All 64 PMM responses were shorter, while 60 pairs simultaneously retained task quality and core action with a shorter response. Figure 2(b) maps paired length reductions across 16 scenarios and three interaction stages. Scenario-level median reductions ranged from 40.1% to 70.5% in Stage 1, 64.6% to 84.3% in Stage 2, and 58.3% to 73.2% in Stage 3, with particularly large reductions in the second stage.

### 5.2.3 Fine-Grained Decision Differences within Shared Action Categories

**Table 5. Fine-grained decision parameters in representative scenarios**

| Scenario | Personality | Persona Prompting | PMM |
|---|---|---|---|
| S01: Project extension | P1 | Request an approximately four-hour extension and assign recalculation tasks | Request a short extension and organize recalculation |
| S01: Project extension | P2 | Submit a formal request when the trigger is met | Request an extension immediately |
| S01: Project extension | P3 | Request a limited two-hour extension | Request a short two-hour extension |
| S01: Project extension | P4 | Set a two-hour decision checkpoint | Request an extension immediately |
| S13: Bonus allocation | P1 | Approximately 60:40 | 55:45 |
| S13: Bonus allocation | P2 | Approximately 55:45 | 60:40 |
| S13: Bonus allocation | P3 | Approximately 60:40 | 50:50 |
| S13: Bonus allocation | P4 | 50:50 | 60:40 |
| S15: Contract renewal | P1 | Approximately 10% in year 1 and 15% in years 2–3 | 25% across all three years |
| S15: Contract renewal | P2 | 12% initially, rising to 15% if conditions are met | Final discount of 15% |
| S15: Contract renewal | P3 | Approximately 20% in year 1 and 18% in years 2–3 | 15% in year 1 and 20% in years 2–3 |
| S15: Contract renewal | P4 | Little or no discount in year 1; at most approximately 15% thereafter | 15% in year 1 and 25% in years 2–3 |

*Note: S13 ratios give the respective bonus shares allocated to the algorithm development team and the collaborating team; S15 values are discount percentages.*

In the project-extension scenario (S01), both methods required the missing data to be verified and submission risks to be controlled. Under P2 and P4, however, PMM requested an extension immediately, whereas Persona Prompting specified a trigger condition or a decision checkpoint. These were the only two core-action disagreements among the 64 matched pairs. In the bonus-allocation scenario (S13), the four matched pairs differed by 5, 5, 10, and 10 percentage points in the algorithm team's share. Contract renewal (S15) likewise revealed different discount levels and temporal structures of concessions.

Figure 2(d) partitions the 64 pairs by differences in core action, resource allocation, and contract parameters: two pairs differed in core action, four in bonus allocation, and four in pricing or contract terms. The other 54 pairs were not assigned to these difference categories. Figure 2(e) summarizes binary A/C behavioral codes for all 128 personality-conditioned responses. The mean of each code was 0.75 across all method–personality combinations, primarily reflecting scenario-level common action patterns. Allocation ratios, discount levels, and the timing of formal procedures provided a finer-grained basis for comparing behavior.

### 5.2.4 Principal Findings

E01 shows substantial agreement between PMM and conventional persona prompting in social decision quality and core action selection, alongside distinct computational and expressive profiles. PMM used an external persistent psychological representation and greater internal computational resources to produce more concise user-facing responses. Several scenarios also revealed observable differences in decision parameters within otherwise shared action categories. E02 examines how these personality conditions enter internal appraisal and candidate-action scoring.

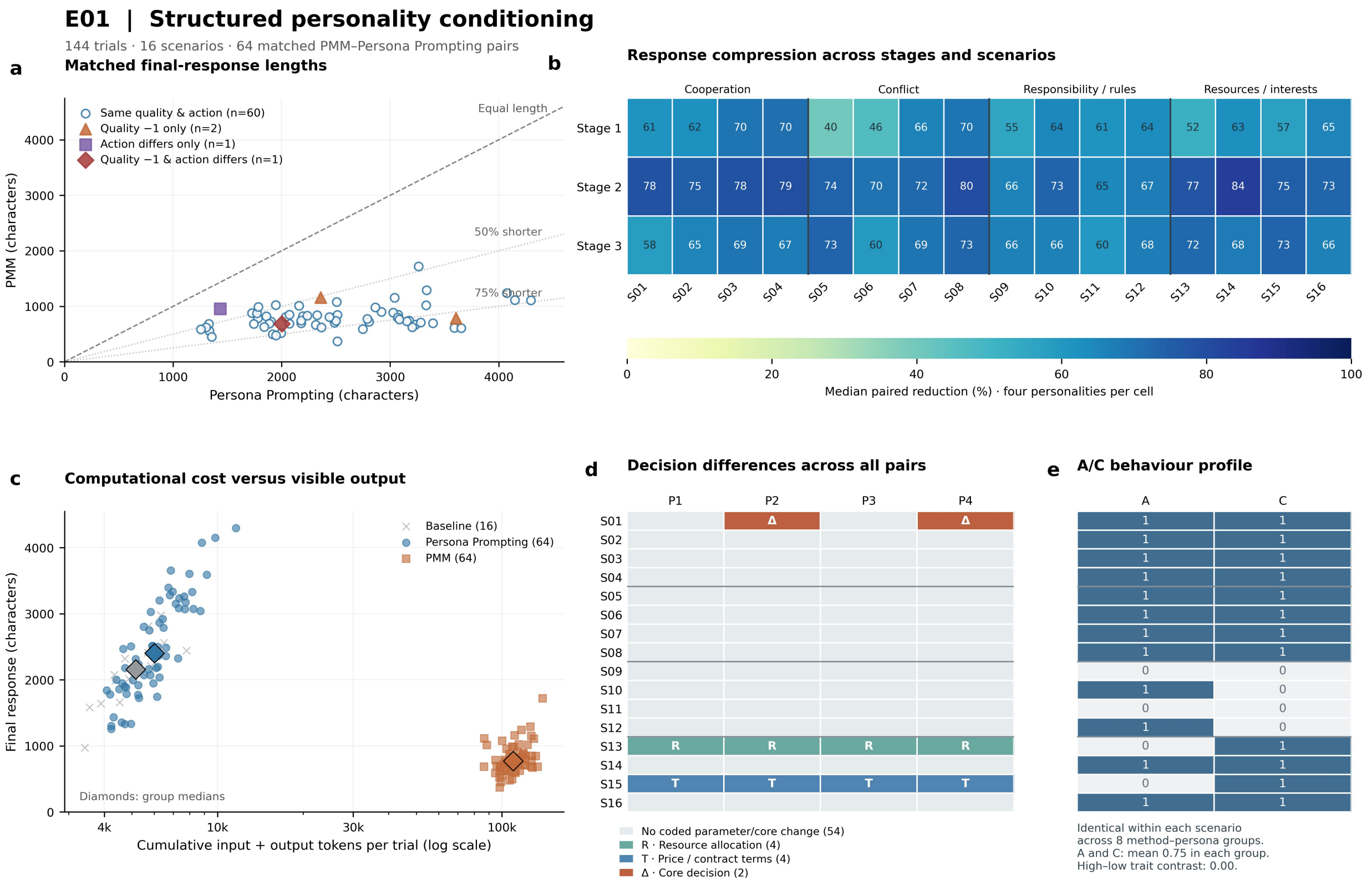

*Figure 2. Social decision-making under structured personality conditioning. (a) Final response lengths for 64 matched PMM–Persona Prompting pairs, with symbols indicating joint task-quality and core-action outcomes; (b) paired length reductions across 16 scenarios and three stages, each cell including four personality configurations; (c) cumulative token consumption and final response lengths for 144 observations, with diamonds indicating group medians; (d) decision difference categories across 64 matched pairs; (e) binary A/C behavioral codes for 128 personality-conditioned observations.*

## 5.3 E02: Personality-Conditioned Cognitive Appraisal and Decision Formation

E02 combined four A/C personality configurations with three repetitions across nine everyday interpersonal scenarios. Of 108 planned trials, 100 yielded complete three-stage trajectories, representing 300 stage-level observations. Nineteen scenario-by-repetition blocks contained all four personality configurations, enabling matched high–low trait comparisons. The stages comprised initial appraisal (Stage 1), contextual reappraisal after new information (Stage 2), and action formation (Stage 3).

### 5.3.1 Overall Cognitive Processing and Social Decision Quality

**Table 6. Cognitive and decision quality across personality configurations**

| Personality | n | Initial understanding | Information integration | Action appropriateness | Cognitive coherence | Total |
|---|---|---|---|---|---|---|
| P1 | 22 | 3.000 | 2.955 | 2.955 | 3.000 | 11.909 |
| P2 | 25 | 3.000 | 3.000 | 3.000 | 3.000 | 12.000 |
| P3 | 27 | 2.815 | 2.963 | 3.000 | 3.000 | 11.778 |
| P4 | 26 | 2.885 | 3.000 | 2.962 | 2.962 | 11.808 |
| Overall | 100 | 2.920 | 2.980 | 2.980 | 2.990 | 11.870 |
| Standard deviation | — | 0.307 | 0.141 | 0.141 | 0.100 | 0.393 |

*Note: Each item has a maximum score of 3; the overall maximum is 12.*

The four profiles achieved near-ceiling task-quality ratings, with a pooled mean of 11.870/12 across 100 complete trials. Overall information integration, action appropriateness, and cross-stage coherence averaged 2.980, 2.980, and 2.990, respectively. The results show consistently high-quality interpretation, evidence integration, and action formation across the personality conditions.

### 5.3.2 Matched Trait Contrasts in Cognitive Appraisal

**Table 7. Matched cognitive-appraisal contrasts across personality conditions**

| Trait dimension | Stage | Threat-appraisal contrast | Clarification-need contrast | Support-orientation contrast |
|---|---|---|---|---|
| C | 1 | −0.01546 | +0.07142 | +0.00934 |
| C | 2 | −0.01157 | +0.05760 | +0.00816 |
| C | 3 | +0.00062 | +0.02191 | +0.01329 |
| A | 1 | −0.01819 | −0.01621 | +0.00566 |
| A | 2 | −0.01984 | +0.01004 | −0.02211 |
| A | 3 | −0.00490 | −0.00571 | +0.00013 |

*Note: High-minus-low trait contrasts within 19 complete four-profile matched blocks at each stage.*

In the 19 complete matched blocks, higher Conscientiousness was associated with greater clarification need, with mean contrasts of +0.07142, +0.05760, and +0.02191 across the three stages. The contrast narrowed as the event became clearer. Conscientiousness-related threat contrasts were negative during the first two stages and approached zero in Stage 3. Agreeableness-related threat contrasts remained negative across stages, while contrasts in clarification and support varied. The association between stable personality conditions and appraisal therefore depended on the cognitive variable and stage of the interaction.

### 5.3.3 Target Trait Contributions and Composite Decision Scores

**Table 8. Matched target contributions, cross-trait contributions, and composite-score contrasts**

| Trait–decision relation | Stage | Target-contribution contrast | Cross-trait contrast | Composite-score contrast | 95% CI (composite) | Aligned blocks |
|---|---|---|---|---|---|---|
| C → Clarification | 1 | +0.08160 | −0.00082 | +0.06204 | [+0.04305, +0.08095] | 17/19 |
| C → Clarification | 2 | +0.05775 | +0.00131 | +0.05124 | [+0.02719, +0.07682] | 15/19 |
| C → Clarification | 3 | +0.02944 | +0.00199 | +0.02026 | [−0.00685, +0.04914] | 12/19 |
| A → Disagreement | 1 | −0.02373 | +0.00168 | −0.05628 | [−0.08630, −0.02521] | 15/19 |
| A → Disagreement | 2 | −0.00608 | +0.00080 | −0.02897 | [−0.05287, −0.00611] | 17/19 |
| A → Disagreement | 3 | −0.00617 | +0.00054 | −0.02182 | [−0.05269, +0.00907] | 14/19 |

*Note: Confidence intervals are based on 20,000 bootstrap resamples of complete blocks.*

The target Conscientiousness contribution to clarify was +0.08160, +0.05775, and +0.02944 across successive stages; all 57 stage-level matched contrasts were positive. The target Agreeableness contribution to disagree was −0.02373, −0.00608, and −0.00617; all 44 nonzero contrasts were negative, with the remaining 13 equal to zero. Cross-trait contrasts—Agreeableness for clarification and Conscientiousness for disagreement—were small, separating target-specific contributions from other trait effects.

Composite clarification-score contrasts were +0.06204, +0.05124, and +0.02026 across stages, while composite disagreement-score contrasts were −0.05628, −0.02897, and −0.02182. Confidence intervals for both composite contrasts excluded zero in the first two stages and narrowed by the final stage. Figure 3(a) displays the distribution of target and cross-trait contributions across the 19 blocks. Figure 3(b) jointly plots the contribution contrasts and composite-score contrasts, revealing general directional alignment alongside blocks where other contextual factors altered the overall action evaluation.

### 5.3.4 Cognitive States during Contextual Updating

**Table 9. Personality contributions and cognitive appraisal across three stages**

| Metric | Stage 1 | Stage 2 | Stage 3 | Reduction (%) |
|---|---|---|---|---|
| Mean absolute C contribution | 0.03806 | 0.02879 | 0.01428 | 62.5 |
| Mean absolute A contribution | 0.01155 | 0.00322 | 0.00306 | 73.5 |
| Threat appraisal | 0.10691 | 0.03030 | 0.02839 | 73.4 |
| Clarification need | 0.35173 | 0.26581 | 0.13184 | 62.5 |
| Support orientation | 0.37240 | 0.35625 | 0.35070 | 5.8 |

**Table 10. Changes in cognitive appraisal from the initial to the final stage**

| Variable | Mean change | 95% Bootstrap CI | Trials with decrease |
|---|---|---|---|
| Threat appraisal | −0.0785 | [−0.097, −0.061] | 69/100 |
| Clarification need | −0.2199 | [−0.260, −0.182] | 94/100 |
| Support orientation | −0.0217 | [−0.035, −0.009] | 55/100 |

*Note: Changes represent Stage 3 minus Stage 1; bootstrap confidence intervals are obtained by resampling complete trajectories.*

With personality conditions held fixed, mean absolute Conscientiousness- and Agreeableness-related contributions decreased by 62.5% and 73.5%, respectively, from Stage 1 to Stage 3. Threat appraisal decreased by 73.4%, clarification need by 62.5%, and support orientation by 5.8%. Clarification need declined in 94 of 100 complete trials (mean change −0.2199; 95% bootstrap CI [−0.260, −0.182]); threat appraisal declined in 69 trials (mean change −0.0785).

Figure 3(c) illustrates scenario-specific changes. In L01, clarification need decreased by 0.3232 and 0.2965 during successive information updates as the reason for a withdrawn invitation became clearer. Threat appraisal in L04 and L06 declined primarily after the first update. In L07, clarification need increased by 0.2394 before declining by 0.3725, demonstrating a non-monotonic trajectory. As information accumulated, the internal action categories shifted from 72 reply, 26 unknown, and two clarify decisions in Stage 1 to 100 reply decisions at each of Stages 2 and 3.

### 5.3.5 Observable Social Actions and Internal Decision Scores

**Table 11. Observable trait-related behavioral scores across personality profiles**

| Personality | n | A-related behavioral score | C-related behavioral score |
|---|---|---|---|
| P1 | 22 | 0.955 | 0.136 |
| P2 | 25 | 1.000 | 0.120 |
| P3 | 27 | 0.889 | 0.185 |
| P4 | 26 | 0.885 | 0.115 |

*Note: Behavioral scores range from −2 to +2.*

**Table 12. Observed social-action distributions in representative scenarios**

| Scenario | Observed action | P1 | P2 | P3 | P4 | Total |
|---|---|---|---|---|---|---|
| L05 | Defer reciprocal commitment | 1 | 2 | 1 | 3 | 7 |
| L05 | Raise commitment later | 2 | 0 | 2 | 0 | 4 |
| L05 | Set future reciprocity boundaries | 0 | 1 | 0 | 0 | 1 |
| L08 | Decline invitation | 1 | 1 | 0 | 0 | 2 |
| L08 | Attend briefly | 1 | 2 | 2 | 3 | 8 |
| L08 | Attend conditionally | 0 | 0 | 1 | 0 | 1 |
| L09 | Request advance notification | 2 | 1 | 2 | 3 | 8 |

| Scenario | Observed action | P1 | P2 | P3 | P4 | Total |
|---|---|---|---|---|---|---|
| L09 | Set waiting-time limit | 0 | 1 | 1 | 0 | 2 |
| L09 | Request a specific improvement plan | 0 | 1 | 0 | 0 | 1 |

Between-profile differences in observable A/C behavioral scores were smaller than those in target trait contributions. Mean A-related scores were 0.979 and 0.887 for the high- and low-Agreeableness groups; the corresponding Conscientiousness-related scores were 0.163 and 0.118. Fourteen of the 19 complete matched blocks retained the same primary action strategy, while five differed, concentrated in L05, L08, and L09. In the first L08 repetition, the four profiles respectively declined an unexpected dinner invitation while suggesting another meeting, maintained a planned rest period, attended briefly with conditions, and attended briefly to assess a professional opportunity. Figure 3(d) juxtaposes composite-score and observable-behavior contrasts; Figure 3(e) maps the 100 completed actions across nine scenarios.

Internal appraisal changes can also be traced to individual conversations. In E02_P1_L01_R1, clarification need progressed from 0.9810 to 0.5995 and 0.2725 as the reason for a friend's withdrawn invitation became clearer; the final response shifted from considering multiple explanations to offering support and allowing rescheduling. In L09, learning that repeated lateness coincided with a job transition reduced attribution of hostile intent, while maintaining requests for advance notification and waiting-time boundaries. These cases show how local reappraisal can coexist with continuity in behavioral principles.

Execution logs recorded 50 ok and 250 failed Reasoner calls across 300 stages. Personality contributions were primarily computed through the RegistryPMM local deterministic fallback path. The matched contrasts therefore characterize the structured trait contributions and their effects on composite scores within that executed pathway.

### 5.3.6 Principal Findings

E02 demonstrates a layered relationship among enduring personality conditions, dynamic cognitive appraisal, local candidate-action contributions, and composite action evaluation. Target trait contributions maintained distinct directions within matched blocks, whereas composite scores responded to accumulating contextual information. Observable action differences were concentrated in interpersonal situations offering several acceptable responses. The separation of internal scoring from final social action provides a computational basis for analyzing personality-conditioned decisions.

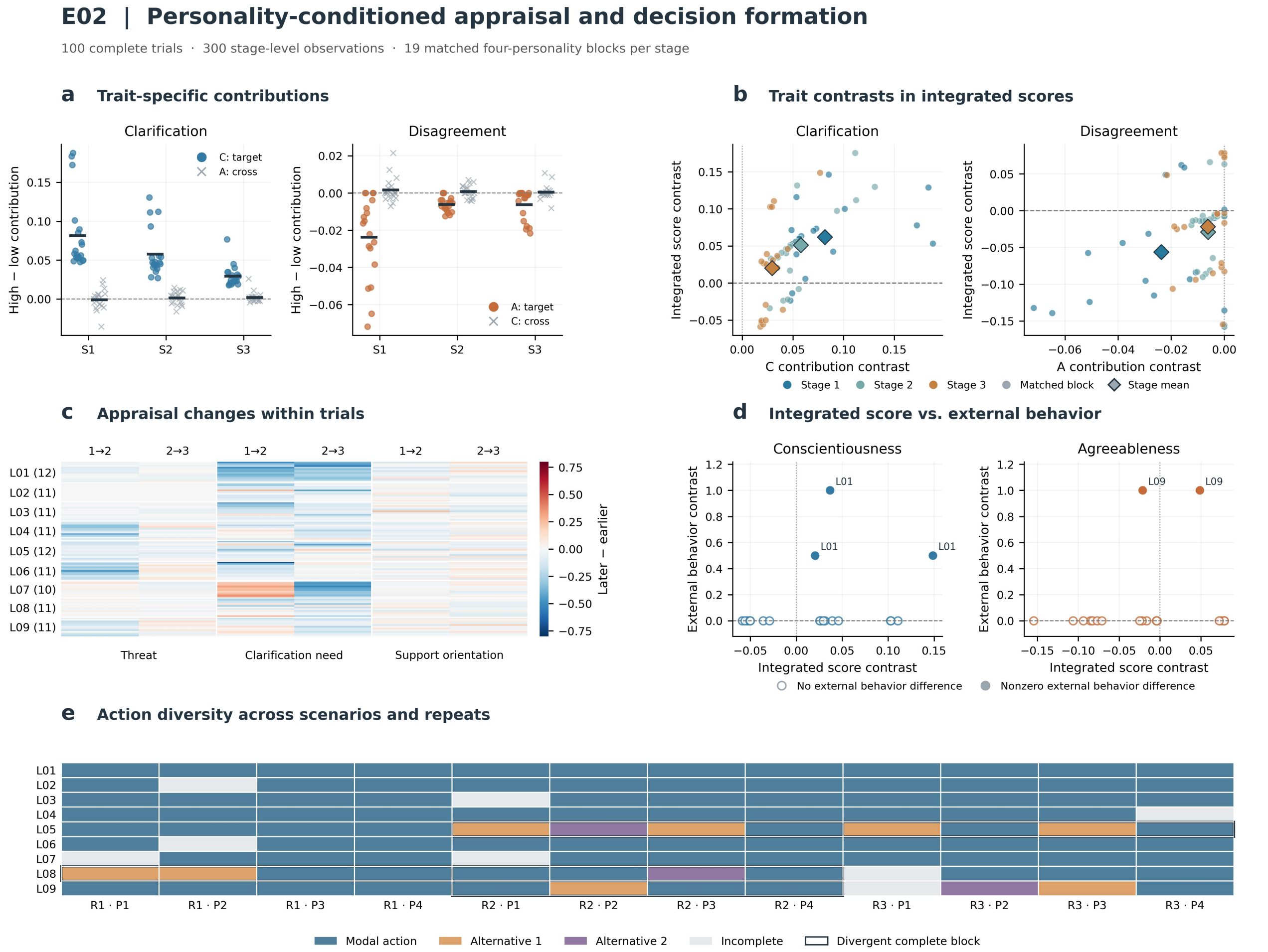

*Figure 3. Personality-conditioned cognitive appraisal and decision formation. S1, initial appraisal; S2, contextual reappraisal following new evidence; S3, action formation. (a) Target and cross-trait contribution contrasts across 19 matched blocks; (b) joint distribution of personality contributions and composite decision-score contrasts; (c) within-trial appraisal changes across 100 completed trajectories; (d) matched contrasts in final composite scores and externally coded behaviors; (e) action distribution across nine scenarios, with black outlines indicating differentiated actions in complete matched blocks.*

## 5.4 E03: Structured Affect Conditioning and Social Behavior Regulation

E03 evaluated positive, neutral, and negative affective conditions under fixed personality profiles in four social scenarios with three repetitions. Of 144 planned records, 130 were complete, yielding 35 triplets with all three affective conditions represented. The analysis distinguished assigned affect, effective runtime states, cognitive appraisal, affective contributions, action scores, and final expression plans.

### 5.4.1 Social Response Quality across Affective Conditions

**Table 13. Affective conditions and overall response appropriateness**

| Metric | Negative | Neutral | Positive |
|---|---|---|---|
| Assigned valence | −0.60 | 0.00 | +0.70 |
| Assigned arousal | 0.55 | 0.35 | 0.60 |
| Assigned dominance | 0.35 | 0.50 | 0.60 |
| Assigned stress | 0.70 | 0.20 | 0.10 |
| Assigned engagement | 0.35 | 0.60 | 0.80 |
| Complete trials | 42 | 43 | 45 |
| Mean appropriateness (0–3) | 2.976 | 2.953 | 3.000 |
| Standard deviation | 0.154 | 0.213 | 0.000 |
| Full-score rate (%) | 97.62 | 95.35 | 100.00 |

*Note: Assigned affective parameters and effective runtime affective states were evaluated separately.*

Mean response appropriateness across the 130 complete records was 2.977/3, with 127 receiving the maximum score. Negative, Neutral, and Positive conditions averaged 2.976, 2.953, and 3.000, respectively. All three conditions produced contextually appropriate social responses. Valence, arousal, dominance, stress, and engagement in Table 13 represent assigned experimental parameters, which are compared below with the states computed during execution.

### 5.4.2 Levels and Distribution of Behavioral Differences

**Table 14. Levels of behavioral differences across matched affective conditions**

| Level | Matched blocks | Proportion (%) |
|---|---|---|
| 0: No substantive difference | 14 | 40.00 |
| 1: Expression and communication | 13 | 37.14 |
| 2: Appraisal or boundary intensity | 7 | 20.00 |
| 3: Response strategy and consequences | 1 | 2.86 |
| Total | 35 | 100.00 |

Twenty-one of 35 complete matched triplets exhibited discernible differences: 13 primarily concerned expression and communication, seven involved event appraisal or interpersonal boundary strength, and one extended to response strategy or behavioral consequences. All 35 triplets retained the same core social-action category. Four triplets showed expressive variation aligned with the assigned positive–negative affective direction, five displayed mixed ordering, and the remaining 26 lacked a clear monotonic ordering. Figure 4(a) maps the depth, direction, and distribution of differences across matched blocks.

### 5.4.3 Contextual Heterogeneity and Representative Behavior

**Table 15. Scenario-level distribution of affect-related behavioral differences**

| Scenario | n | Level 0 | Level 1 | Level 2 | Level 3 | Any difference (%) |
|---|---|---|---|---|---|---|
| L01: Withdrawn invitation | 9 | 2 | 6 | 1 | 0 | 77.78 |
| L04: Misinterpreted remark | 8 | 6 | 2 | 0 | 0 | 25.00 |
| L07: Offensive joke | 9 | 5 | 4 | 0 | 0 | 44.44 |
| L09: Repeated lateness | 9 | 1 | 1 | 6 | 1 | 88.89 |
| Total | 35 | 14 | 13 | 7 | 1 | 60.00 |

Differences were concentrated in repeated lateness (L09): eight of nine matched triplets differed at some level, and seven involved appraisal, interpersonal boundaries, or strategy. In contrast, misinterpreted remarks (L04) largely elicited clarification and repair, offensive jokes (L07) elicited apology and boundary respect, and the withdrawn-invitation scenario (L01) differed mainly in expressions of concern, inquiries, and future contact arrangements.

For example, in a P4 repetition of L09, the Negative condition stated that continued lateness would prompt reconsideration of the relationship, whereas the Positive condition requested advance notification to facilitate scheduling. Both responses favored communication but differed in the consequences attached to repeated violations and the strength of relational boundaries.

### 5.4.4 Effective Affect, Cognitive Appraisal, and Composite Decisions

**Table 16. Paired contrasts in affective states, appraisals, and decision scores**

| Variable (Negative – Positive) | Stage 1 | Stage 2 | Stage 3 |
|---|---|---|---|
| Computed valence | −0.00952 | +0.00437 | +0.04252 |
| Threat appraisal | +0.04068 | +0.00313 | +0.01031 |
| Clarification need | +0.01582 | +0.03077 | +0.00341 |
| Support orientation | +0.00329 | +0.02171 | +0.00167 |
| Affect contribution to reply | −0.00061 | +0.00047 | +0.00266 |
| Affect contribution to clarification | +0.00015 | +0.00027 | +0.00001 |
| Composite reply score | −0.03938 | +0.24214 | +0.12555 |
| Composite clarification score | −0.00457 | −0.00887 | −0.02487 |

*Note: Each stage is based on 35 complete matched affective triplets.*

Figure 4(b) presents Negative-minus-Positive paired contrasts in computed valence and cognitive appraisal. The valence contrast changed from −0.00952 at Stage 1 to +0.04252 at Stage 3. Final-stage effective valence averaged 0.07533, 0.06207, and 0.03281 under Negative, Neutral, and Positive conditions, respectively, an ordering different from the assigned conditions. Among the 130 complete records, 119 had first-stage effective affect fields identical to the default snapshot. The four-dimensional runtime-state RMSE relative to assigned vectors was 0.3081, 0.3117, and 0.3218 across successive stages, compared with 0.0058, 0.0164, and 0.0367 relative to the default vector.

Cognitive and decision variables exhibited distinct temporal distributions. The Negative-minus-Positive threat-appraisal contrast was +0.04068 in Stage 1 and narrowed to +0.00313 after new information. The larger clarification-need contrast occurred in Stage 2. Figure 4(c) places the affective contribution to reply alongside composite reply-score contrasts. Their mean absolute paired differences were 0.00119, 0.00223, and 0.00485 for affective contributions and 0.23365, 0.42716, and 0.22419 for composite scores. Local affective contributions and composite decision scores thus differed both in magnitude and in the stage at which variation peaked.

### 5.4.5 Expression Planning and Interaction Organization

**Table 17. Final expression-plan distribution across affective conditions**

| Expression plan | Negative | Neutral | Positive |
|---|---|---|---|
| Minimal Reply | 28 | 26 | 34 |
| Respond and Yield | 7 | 9 | 1 |
| Total | 35 | 35 | 35 |
| Respond and Yield (%) | 20.00 | 25.71 | 2.86 |

Respond and Yield appeared in seven Negative, nine Neutral, and one Positive matched trials. The paired binary comparison yielded Cochran's $Q = 6.933$ ($df = 2$, $p = 0.031$). Figure 4(d) identifies 20 blocks that used Minimal Reply in every condition and 15 blocks with condition-dependent expression plans: seven used Respond and Yield only under Neutral, six only under Negative, and two used it under two-condition combinations. Structured expression planning thus captured an aspect of interaction organization separate from core action selection.

### 5.4.6 Principal Findings

E03 reveals affect-related variables in distinct runtime representations, local action contributions, and expression plans. Observable differences centered on communication style, event appraisal, and interpersonal boundaries. The contrast between assigned affective conditions and effective runtime states further identifies the numerical characteristics of affect transmission, making input configuration, internal computation, and behavioral expression separately observable.

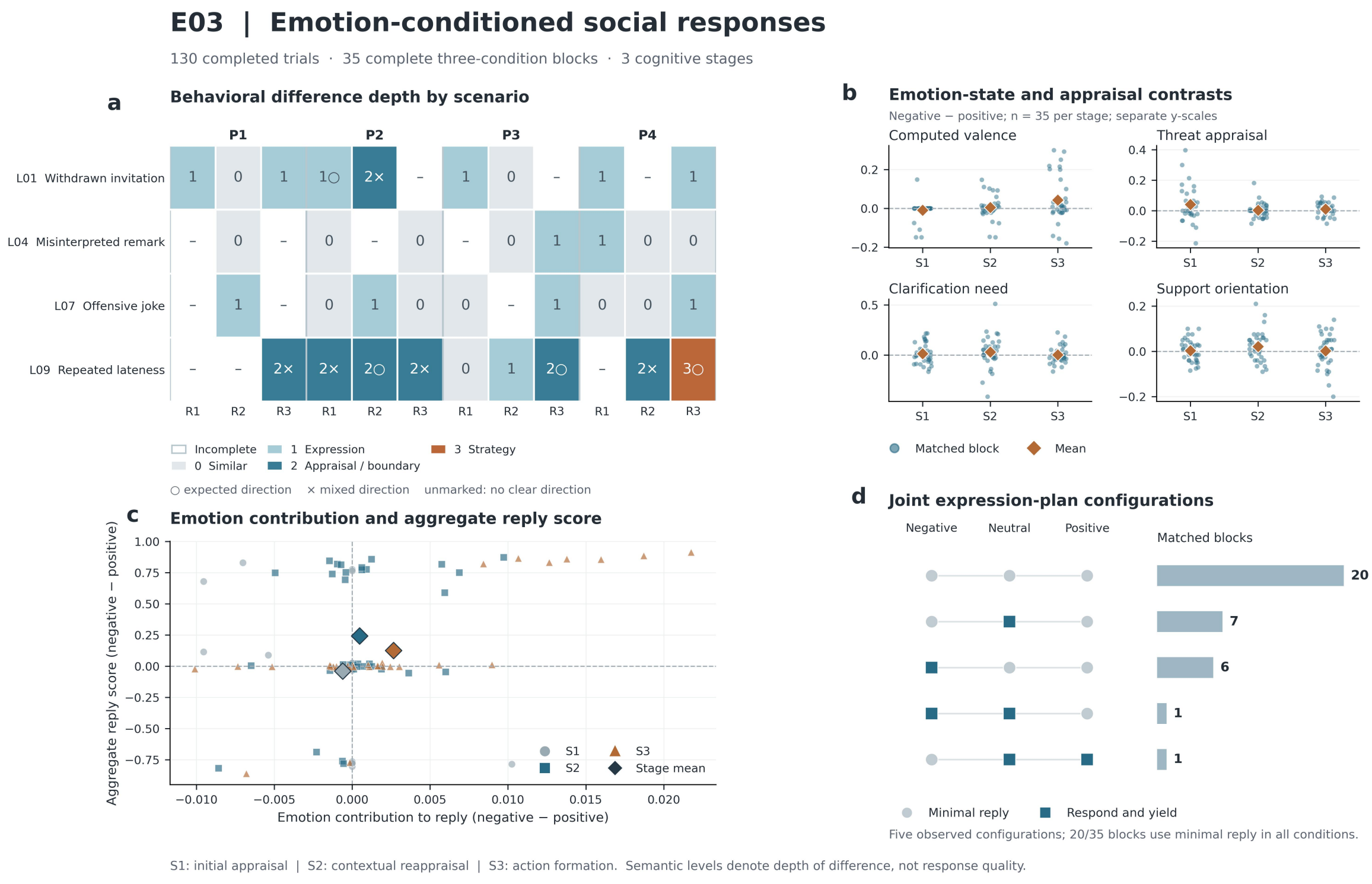

*Figure 4. Emotion-conditioned cognitive appraisal, decisions, and expression plans. (a) Levels and directions of behavioral differences across 35 matched triplets; (b) Negative-minus-Positive contrasts in effective valence and cognitive appraisal; (c) joint distribution of affective contributions to reply and composite reply-score contrasts; (d) five observed expression-plan configurations across the three affective conditions.*

## 5.5 E04: Record-Grounded Self-Evaluation and Reflection

E04 examined how an artificial individual evaluates a record of its own earlier predictions after receiving additional contextual information. Across nine scenarios, four personality configurations, and three repetitions, 108 sequences were initiated and 83 contained a complete prediction, three-stage action, and self-evaluation record. The prediction was made under the initial information; the final action followed subsequent evidence. The analyses combine archived rubric scores, full-sample lexical comparisons, and independently adjudicated cases, with the five panels of Figure 5 displaying these different sources of evidence.

### 5.5.1 Rubric-Based Prediction and Self-Evaluation Ratings

**Table 18. Overall prediction and record-grounded self-evaluation ratings**

| Evaluation dimension | Scale | Mean | Standard deviation | Full-score trials (n/83) |
|---|---|---|---|---|
| Prediction alignment | 0–2 | 1.988 | 0.110 | 82 (98.80%) |
| Discrepancy recognition | 0–3 | 2.916 | 0.280 | 76 (91.57%) |
| Source fidelity | 0–3 | 2.976 | 0.154 | 81 (97.59%) |
| Reason explanation | 0–3 | 2.265 | 0.444 | 22 (26.51%) |
| Total | 0–11 | 10.145 | 0.608 | 21 (25.30%) |

Across the 83 complete sequences, mean prediction alignment was 1.988/2, discrepancy recognition 2.916/3, source fidelity 2.976/3, and reason explanation 2.265/3, producing a mean total of 10.145/11 (Table 18). Full marks were assigned in 82, 76, 81, and 22 records on the four respective dimensions. Fifty-three records received maximum scores for the first three dimensions and a score of 2 for explanation; another 21 received the maximum on all four dimensions. These are distributions of archived rubric-based ratings. They distinguish the high frequency of favorable discrepancy and fidelity ratings from the smaller share of maximum explanation scores, while the independent semantic reference set is reported separately below.

### 5.5.2 Prediction–Action Relationships and Self-Descriptions

**Table 19. Self-evaluation scores by prediction–action relationship**

| Decision relationship | n (%) | Prediction alignment | Discrepancy recognition | Source fidelity | Reason explanation |
|---|---|---|---|---|---|
| Core choice maintained | 30 (36.14) | 2.000 | 2.833 | 2.967 | 2.033 |
| Implementation adjusted | 40 (48.19) | 1.975 | 2.950 | 2.975 | 2.200 |
| Core choice changed | 13 (15.66) | 2.000 | 3.000 | 3.000 | 3.000 |

In the archived behavioral classification, 30 of 83 complete sequences maintained the predicted core choice, 40 adjusted its implementation, and 13 changed the core decision (Table 19). The 53 adjusted or changed sequences accounted for 63.86% of complete records. All 13 classified core-choice changes were concentrated in L05 (six records) and L08 (seven records). The additional information introduced during the action stages therefore often altered how a choice was implemented, and sometimes changed the choice itself. These whole-sample categories describe the archived classification; the independently reviewed reference cases for L08 are examined in Section 5.5.3.

### 5.5.3 Independently Adjudicated Decisions and Record Fidelity

The retrospective self-descriptions labeled 55 sequences as broadly consistent, 18 as partially consistent, and three as changed; seven did not provide an explicit overall label. Among the 40 implementation adjustments, 28 were self-described as consistent. Among the 13 classified core-choice changes, three were described as consistent, eight as partially consistent, and two as changed. These self-description labels refer to perceived continuity of guiding principles, which need not match the coding of individual actions. They are therefore reported separately from independent judgments of whether a concrete decision change was correctly recognized.

Ten complete L08 sequences received independent manual adjudication of the predicted choice, enacted choice, and self-evaluation. Eight predicted refusal and two anticipated conditional attendance; the final actions comprised nine brief or conditional attendances and one refusal. Seven sequences involved a verified refusal-to-attendance change, and all seven self-evaluations explicitly recognized that reversal. The remaining two conditional-attendance predictions were compatible with their enacted actions, and one refusal remained unchanged; their self-evaluations also distinguished these outcomes. Figure 5(c) displays the ten decision paths and checks. For example, in E04_P1_L08_R1, the individual initially anticipated remaining home, then chose brief attendance after learning of a possible professional connection while retaining a personal time boundary in its explanation.

The examined L05 case E04_P1_L05_R1 provides a second example of context-sensitive reinterpretation: after learning of a family hospitalization, the agent shifted from asking about an unfulfilled reciprocal meal to offering help, and its self-evaluation identified the new information as relevant to this changed priority. The evidence audit also located two concrete reconstruction errors. In E04_P3_L02_R1, the self-evaluation misstated the anticipated afternoon report arrangement; in E04_P3_L04_R3, it confused a private communication with a public response. Figure 5(e) contrasts these cases with the successful L08 and L05 interpretations, showing why recognizing a decision difference and faithfully reconstructing its supporting record are separate evaluation targets.

### 5.5.4 Full-Sample Lexical Divergence across Stages

For the 83 complete sequences, prediction-to-Stage 1 and prediction-to-Stage 3 cosine similarities averaged 0.04749 and 0.02748, respectively, in a shared character 2–4-gram TF–IDF representation. The mean paired change (Stage 3 minus Stage 1) was −0.02001, with a trial-level bootstrap 95% confidence interval of [−0.02751, −0.01235]. In 62 of 83 sequences, the final response had lower lexical similarity to the prediction than the Stage 1 response. Figure 5(a) maps these paired differences by scenario and personality–repetition cell, with the manually adjudicated L08 cases marked separately; Figure 5(b) shows the paired similarity coordinates.

The scenario-level distributions in Figure 5(d) indicate that the magnitude and direction of lexical divergence varied by context. These statistics quantify textual change across information stages, including changes in response length and phrasing. They do not measure whether a core action changed or whether a retrospective explanation was correct. Their value is to characterize all available records and identify where independent semantic examination can most directly test action-change recognition.

### 5.5.5 Self-Evaluation across Personality Profiles

**Table 20. Self-evaluation performance across personality conditions**

| Personality | n | Prediction alignment | Discrepancy recognition | Source fidelity | Reason explanation | Total (0–11) |
|---|---|---|---|---|---|---|
| P1 (high A, high C) | 17 | 2.000 | 2.941 | 3.000 | 2.235 | 10.176 |
| P2 (high A, low C) | 22 | 2.000 | 2.818 | 3.000 | 2.273 | 10.091 |
| P3 (low A, high C) | 25 | 2.000 | 2.920 | 2.960 | 2.240 | 10.120 |
| P4 (low A, low C) | 19 | 1.947 | 3.000 | 2.947 | 2.316 | 10.211 |

Descriptive rubric totals ranged from 10.091 to 10.211 across the four personality configurations (Table 20). Prediction alignment, discrepancy recognition, and source fidelity received high mean ratings in every configuration, while scores for explanation were more variable. The four profiles contained three, four, three, and three core-choice changes, respectively, in the archived classification. Together, the profile summaries show that the same prediction–action–evaluation procedure was applied across different A/C conditions; the independent reference judgments in Figure 5(c) cover the L08 subset rather than all 83 records.

### 5.5.6 Principal Findings

E04 demonstrates a record-grounded form of behavioral self-monitoring: earlier predictions and later actions can be compared, their differences described, and explanations tied to newly available information. In ten independently adjudicated L08 sequences, the agent correctly identified all seven verified core-choice reversals while distinguishing two compatible choices and one maintained refusal. Across the 83 complete sequences, archived rubric ratings and lexical divergence provide complementary descriptions of self-evaluation and response change; the localized reconstruction errors show that source fidelity requires explicit checking even when an explanation is coherent. Figure 5 integrates the full-record distribution with adjudicated decisions and auditable case evidence.

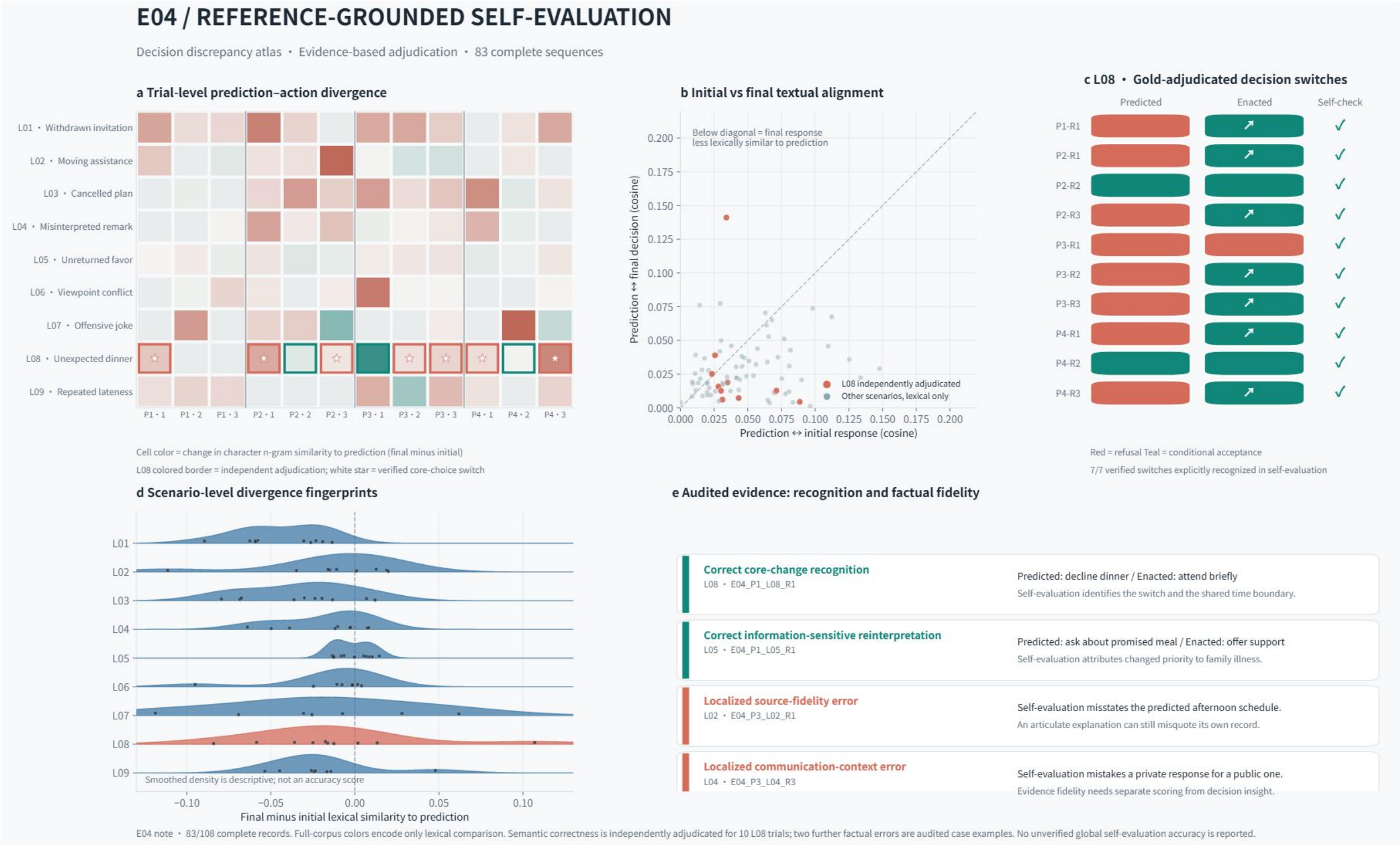

Figure 5. Record-grounded self-evaluation with full-sample lexical comparisons and independently adjudicated cases. (a) Trial-level change in prediction-to-response character n-gram similarity between the initial and final action stages across 83 complete sequences; outlines and stars identify the L08 manually reviewed cases and verified core-choice changes. (b) Prediction-to-Stage 1 versus prediction-to-Stage 3 TF–IDF cosine similarity, highlighting the independently reviewed L08 subset. (c) Predicted and enacted action categories in ten manually adjudicated L08 sequences, including seven correctly recognized refusal-to-attendance transitions. (d) Scenario-level distributions of lexical similarity change. (e) Two illustrative successful evaluations and two localized factual-reconstruction errors. Lexical divergence is not a measure of behavioral or self-evaluation accuracy; independent semantic adjudication covers the ten L08 sequences.

## 5.6 E05: From Experiential Adaptation to Long-Term Personality Development

E05 combined four initial personality profiles (P1–P4), three experience environments (E1–E3), and four learning configurations (C1–C4), yielding 48 longitudinal trajectories. Each trajectory contained 12 social events across four developmental waves (W1–W4), spanning 36 simulated days. Initial high and low A/C values were 0.80 and 0.20; personality updates aggregated evidence across events and used a bounded step of at most 0.03 per wave. Outcomes included enduring traits, experiential trust and caution, continuous behavioral ratings, and categorical actions.

**Table 21. Longitudinal experience environments and learning conditions**

| Factor | Definition |
|---|---|
| E1: Cooperative mastery | Sustained cooperation, reciprocity, and mastery-related feedback |
| E2: Betrayal and accountability | Repeated betrayal, responsibility conflicts, and accountability demands |
| E3: Transient disruption | Transient negative shocks followed by recovery events |
| C1: Full PMM | Joint learning of short-term affect, experiential cognitive states, and enduring personality |
| C2: Memory/cognition-only learning | Short-term affect and cognitive-state updating with enduring traits frozen |
| C3: Static PMM | Initial personality and short-term affect retained; experiential states and enduring traits frozen |
| C4: Prompt + memory | Baseline combining natural-language persona descriptions with conversational memory |

### 5.6.1 Experience-Dependent Trait Development and Individual Continuity

**Table 22. Initial and final trait states under C1**

| Environment | Personality | W0 (A, C) | W4 (A, C) | ΔA | ΔC |
|---|---|---|---|---|---|
| E1 | P1 | (0.80, 0.80) | (0.88, 0.88) | +0.080 | +0.080 |
| E1 | P2 | (0.80, 0.20) | (0.88, 0.32) | +0.080 | +0.120 |
| E1 | P3 | (0.20, 0.80) | (0.32, 0.88) | +0.120 | +0.080 |
| E1 | P4 | (0.20, 0.20) | (0.32, 0.32) | +0.120 | +0.120 |
| E2 | P1 | (0.80, 0.80) | (0.68, 0.88) | −0.120 | +0.080 |
| E2 | P2 | (0.80, 0.20) | (0.68, 0.32) | −0.120 | +0.120 |
| E2 | P3 | (0.20, 0.80) | (0.12, 0.88) | −0.080 | +0.080 |
| E2 | P4 | (0.20, 0.20) | (0.12, 0.32) | −0.080 | +0.120 |
| E3 | P1 | (0.80, 0.80) | (0.80, 0.80) | 0.000 | 0.000 |
| E3 | P2 | (0.80, 0.20) | (0.80, 0.20) | 0.000 | 0.000 |
| E3 | P3 | (0.20, 0.80) | (0.20, 0.80) | 0.000 | 0.000 |
| E3 | P4 | (0.20, 0.20) | (0.20, 0.20) | 0.000 | 0.000 |

*Note: C1 denotes the full-learning condition; A, Agreeableness; C, Conscientiousness.*

Under full learning (C1), positive cooperation (E1) increased both Agreeableness and Conscientiousness by a mean of +0.100. Betrayal and accountability (E2) decreased Agreeableness by 0.100 and increased Conscientiousness by 0.100; transient disruption (E3) left both enduring traits unchanged. Figure 6(c) traces individual profiles from W0 to W4. In E1, the mean cumulative increases in both traits across successive waves were 0, 0.030, 0.060, 0.085, and 0.100. Agreeableness followed the opposite direction in E2, while Conscientiousness increased.

Across the 12 C1 trajectories, initial-to-final Pearson correlations were 0.9613 for A and 0.9862 for C, with Spearman correlations of 0.8783 and 0.9045. All four high–low profile combinations remained within their initial quadrants. Within each sustained-experience environment, the final high–low separation was 0.56, retaining 93.3% of the initial 0.60 difference. Experience-dependent trait movement thus coexisted with continuity in individual differences under bounded updating.

### 5.6.2 Incremental Behavioral Effects of Trait Learning beyond Memory

**Table 23. Changes in experiential and relational states**

| Environment | Condition | ΔTrust | ΔCaution | ΔMastery |
|---|---|---|---|---|
| E1 | C1/C2 | +0.352 | −0.198 | +0.325 |
| E1 | C3/C4 | 0.000 | 0.000 | 0.000 |
| E2 | C1/C2 | −0.335 | +0.365 | +0.325 |
| E2 | C3/C4 | 0.000 | 0.000 | 0.000 |
| E3 | C1/C2 | +0.005 | +0.016 | 0.000 |
| E3 | C3/C4 | 0.000 | 0.000 | 0.000 |

*Note: Changes are mean values for Event 12 minus Event 1. C1 and C2 share these states in all 144 matched events.*

Figure 6(b) depicts trust, caution, and mastery across 12 events. C1 and C2 had identical values for these variables in all 144 matched events. Under E1, trust increased by 0.352, caution declined by 0.198, and mastery increased by 0.325. Under E2, trust declined by 0.335, caution increased by 0.365, and mastery again increased by 0.325. This matched experiential-state structure isolates the contrast between conditions with and without enduring-trait updating.

**Table 24. Behavioral development across experience environments and learning conditions**

| Environment | Condition | ΔInterpers. | ΔPlanning | ΔRules | ΔConflict | ΔFlexibility | Δa_eff | Δc_eff |
|---|---|---|---|---|---|---|---|---|
| E1 | C1 | +0.458 | +0.375 | +0.208 | +0.500 | +0.333 | +0.144 | +0.103 |
| E1 | C2 | +0.292 | +0.042 | 0.000 | +0.208 | +0.083 | +0.078 | +0.016 |
| E1 | C3 | +0.042 | +0.042 | −0.042 | −0.292 | 0.000 | −0.009 | +0.005 |
| E1 | C4 | +0.542 | −0.208 | −0.292 | +0.042 | +0.250 | +0.117 | −0.016 |
| E2 | C1 | −0.500 | +0.583 | +0.375 | −0.083 | 0.000 | −0.106 | +0.141 |
| E2 | C2 | −0.333 | +0.333 | +0.417 | −0.208 | +0.042 | −0.084 | +0.083 |
| E2 | C3 | −0.125 | +0.083 | +0.083 | +0.167 | −0.083 | −0.003 | +0.006 |
| E2 | C4 | −0.042 | 0.000 | +0.042 | −0.042 | 0.000 | −0.008 | +0.013 |
| E3 | C1 | +0.167 | 0.000 | −0.083 | −0.042 | −0.042 | +0.012 | −0.008 |

| Environment | Condition | ΔInterpers. | ΔPlanning | ΔRules | ΔConflict | ΔFlexibility | Δa_eff | Δc_eff |
|---|---|---|---|---|---|---|---|---|
| E3 | C2 | −0.125 | −0.042 | −0.042 | −0.125 | +0.042 | −0.011 | +0.003 |
| E3 | C3 | −0.333 | 0.000 | −0.083 | −0.125 | −0.250 | −0.026 | +0.004 |
| E3 | C4 | +0.083 | −0.125 | −0.167 | −0.042 | −0.375 | −0.029 | −0.019 |

*Note: Values are mean W4–W1 changes over four matched personality trajectories. Interpers. = interpersonal consideration; Planning = responsibility and planning. Behavioral scores use a 1–5 scale; a_eff and c_eff denote effective behavioral coordinates.*

In E1, interpersonal consideration and responsibility-oriented planning under C1 increased by 0.458 and 0.375 points from W1 to W4. In E2, the corresponding changes were −0.500 and +0.583. Relative to C2, C1 showed additional planning gains of +0.333 in E1 and +0.250 in E2. Across the eight matched initial-profile comparisons in the two sustained environments, seven increments were positive and one was unchanged. Compared with C4, the incremental planning gain under C1 was +0.583 in each environment. By contrast, C4 achieved a greater improvement in interpersonal consideration in E1 (+0.542 versus +0.458), and C2 achieved a slightly greater improvement in rule adherence in E2 (+0.417 versus +0.375).

Figure 6(a) plots matched C1–C2 behavioral changes, while Figure 6(e) shows wave-specific increments relative to W1. In E1, the additional planning change moved from −0.167 at W2 to +0.250 at W3 and +0.333 at W4. The corresponding progression in E2 was +0.167, +0.208, and +0.250. The pattern differed by environment, developmental period, and behavioral dimension, with sustained trait learning particularly reflected in responsibility-oriented planning and context-dependent interpersonal adjustment.

### 5.6.3 Temporal Separation of Affective Recovery and Long-Term Updating

**Table 25. Affective recovery after transient perturbations in E3**

| Condition | Valence before shock | Valence after shock | Valence after recovery | Valence recovery (%) | Stress before shock | Stress after shock | Stress after recovery | Stress recovery (%) |
|---|---|---|---|---|---|---|---|---|
| C1 | 0.040 | −0.912 | −0.227 | 71.97% | 0.503 | 0.980 | 0.655 | 68.17% |
| C2 | 0.041 | −0.915 | −0.213 | 73.43% | 0.508 | 0.980 | 0.671 | 65.50% |
| C3 | 0.031 | −0.916 | −0.226 | 72.94% | 0.511 | 0.980 | 0.659 | 68.32% |
| C4 | −0.333 | −0.942 | −0.694 | 40.69% | 0.807 | 0.980 | 0.967 | 7.72% |

*Note: Each condition includes four perturbation cycles and four personality profiles (16 cycles); recovery percentages are calculated from condition means.*

E3 introduced brief negative perturbations at Events 2, 5, 8, and 11, followed by recovery-related information. Under C1, mean valence fell from 0.040 before perturbation to −0.912 after perturbation, then recovered to −0.227. Stress increased from 0.503 to 0.980 before declining to 0.655. Valence recovery ranged from 71.97% to 73.43% across C1–C3 and stress recovery from 65.50% to 68.32%; the corresponding C4 values were 40.69% and 7.72%. Figure 6(f) presents pre-perturbation, post-perturbation, and recovery states alongside the 16 individual perturbation–recovery cycles per condition.

Meanwhile, C1 retained its initial enduring A/C parameters throughout E3, and mean responsibility-oriented planning was 3.000 at both W1 and W4. Transient affect, intermediate behavioral adjustment, and enduring trait stability therefore evolved along distinct timescales within the same continuous interaction sequences.

### 5.6.4 Action Categories, Strategic Transitions, and Longitudinal Continuity

**Table 26. Cross-wave behavioral continuity and categorical action changes**

| Condition | Interpersonal r | Responsibility r | Trajectories with multiple action categories |
|---|---|---|---|
| C1 | 0.893 | 0.953 | 2/12 |
| C2 | 0.931 | 0.975 | 0/12 |
| C3 | 0.975 | 0.985 | 0/12 |
| C4 | 0.980 | 0.912 | 6/12 |

*Note: The correlations are Pearson r between W1 and W4 across the 12 personality–environment trajectories.*

Figure 6(d) summarizes all 576 categorical actions across 48 trajectories. In E2 under C1, P1 moved from principled cooperation (A1) toward assertive rule enforcement (A3): it first switched at Event 4, briefly returned to A1 at Event 5, and maintained A3 from Events 6–12. P2 moved from flexible accommodation (A2) to self-protective detachment (A4) at Event 5 and maintained A4 thereafter. P3 and P4 retained their initial action categories. All C2 and C3 trajectories preserved their initial categories, whereas six of 12 C4 trajectories displayed category changes; under E3, all 48 C4 actions were classified as A4.

Under C1, W1–W4 correlations in interpersonal consideration and responsibility-oriented planning were 0.893 and 0.953. Taken together, Figures 6(c)–(e) show that relatively high behavioral continuity coexisted with localized strategy shifts and progressive personality changes. Experiential state updating, incremental behavioral effects of enduring-trait learning, and affective recovery operated as distinct update pathways.

### 5.6.5 Principal Findings

E05 shows an incremental behavioral effect of enduring-trait updating in longitudinal comparisons matched by initial personality and experiential environment. Sustained cooperation and accountability yielded distinct A/C development patterns, while learning configurations differed in planning and other behavioral outcomes without erasing the original high–low trait structure. Short-term affect followed a separate recovery trajectory, whereas enduring traits changed gradually with accumulated experience. The six figure panels connect experiential states, personality trajectories, continuous behavior, and categorical actions.

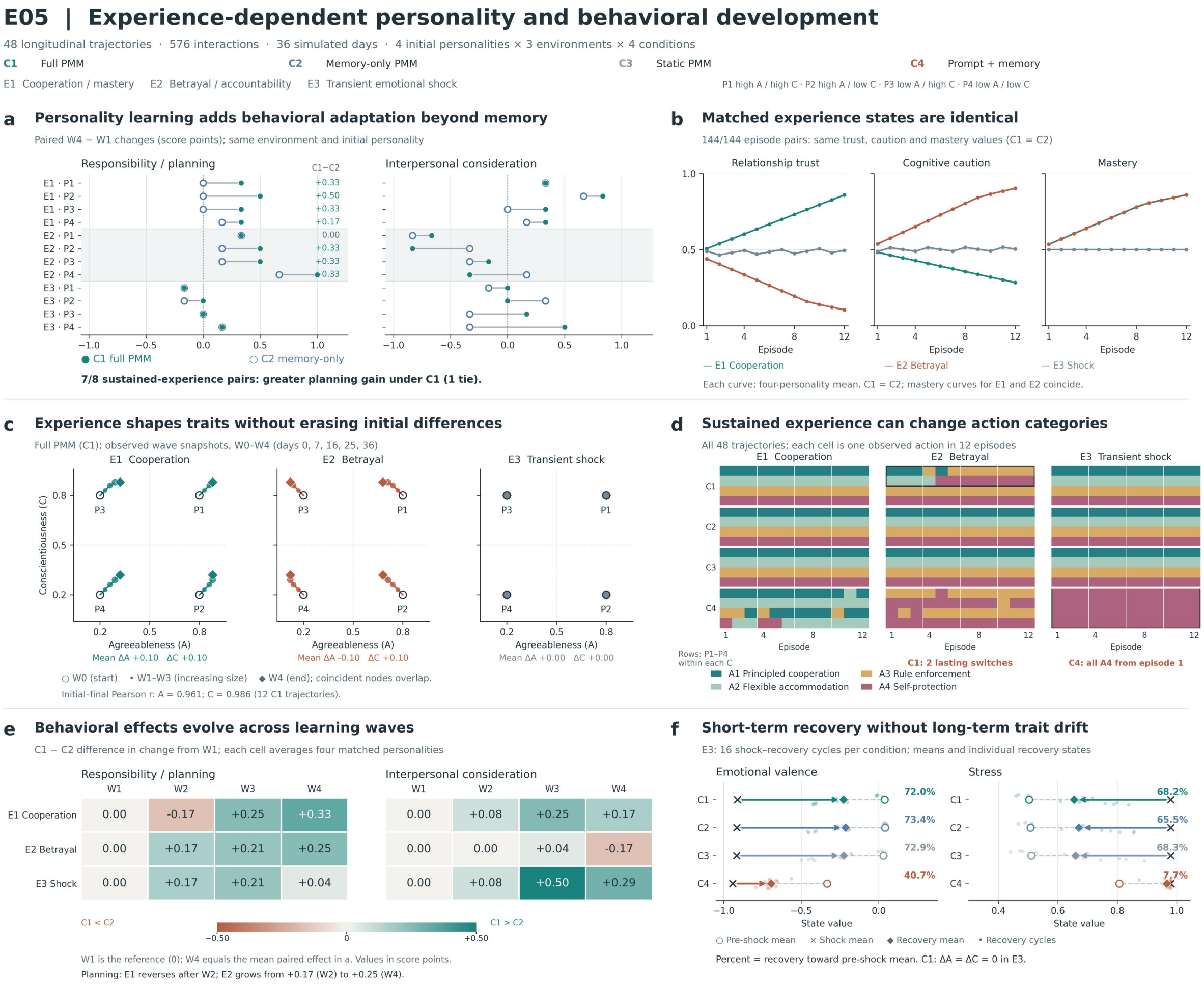

| Responsibility / planning | W1 | W2 | W3 | W4 |
|---|---|---|---|---|
| E1 Cooperation | 0.00 | -0.17 | +0.25 | +0.33 |
| E2 Betrayal | 0.00 | +0.17 | +0.21 | +0.25 |
| E3 Shock | 0.00 | +0.17 | +0.21 | +0.04 |

| Interpersonal consideration | W1 | W2 | W3 | W4 |
|---|---|---|---|---|
| E1 Cooperation | 0.00 | +0.08 | +0.25 | +0.17 |
| E2 Betrayal | 0.00 | 0.00 | +0.04 | -0.17 |
| E3 Shock | 0.00 | +0.08 | +0.50 | +0.29 |

*Figure 6. Experience-dependent personality and behavioral development. (a) Paired W4–W1 behavioral changes for C1 and C2; (b) trust, caution, and mastery trajectories across 144 matched events; (c) enduring-trait trajectories under C1 from W0 to W4; (d) 576 categorical actions across 48 trajectories, highlighting C1 transitions in E2 and uniform C4 actions in E3; (e) C1-minus-C2 incremental behavioral change relative to W1; (f) pre-perturbation, post-perturbation, and recovery states in E3, including recovery observations and percentages from 16 cycles.*

# 6. Discussion

## 6.1 From Personality-like Outputs to Computable Individual Psychological Mechanisms

The central questions in artificial personality research are evolving. Persona prompting and psychometric studies have demonstrated that large language models (LLMs) can express different trait levels in generated text and task responses [1,3]. The more consequential question for persistent artificial personality is how individual differences are represented and maintained, how they shape the interpretation and evaluation of the same event, and how they remain continuous or change through subsequent experience.

PMM addresses this question through a distinct computational organization. Enduring psychological characteristics, transient affective states, situational appraisals, and experiential records are maintained as states with differentiated update conditions. The independent psychological module produces structured decision conditions, which the General LLM subsequently realizes in language. This organization shifts the object of investigation from the final response to the computations that precede it. Personality becomes an enduring individual condition that can contribute to candidate-action evaluation, whose contribution can be traced, and whose parameters can undergo constrained updates as experience accumulates.

This shift has two theoretical implications. Personality continuity concerns the maintenance and use of individual psychological states across situations, extending beyond consistency in conversational style. Personality development concerns the effects of accumulated experience on subsequent interpretations and behavioral trade-offs, extending beyond the retention of conversational history. The five experiments investigate complementary aspects of this process—observable behavior, internal contributions, short-term regulation, self-evaluation, and longitudinal updating—and establish distinct empirical targets for evaluating artificial personality as a computational hypothesis.

### 6.2 Behavioral Convergence and Internal Differentiation: Identifying Personality Contributions

Taken together, E01 and E02 reveal a particularly informative finding: **similar final actions can be supported by different internal evaluation structures.** In E01, PMM and Persona Prompting selected the same core action in 62 of 64 matched cases, with closely comparable task quality. At the level of action categories, the two conditioning approaches were therefore highly similar. E02 extended the observations to candidate-action scores. Across 19 matched four-personality blocks, the conscientiousness-related contribution to clarification consistently favored the high-conscientiousness conditions, while all nonzero agreeableness-related contrasts in disagreement contributions favored the high-agreeableness direction. These contrasts propagated to composite decision scores to varying degrees. Nevertheless, 14 of the 19 blocks retained the same primary social action across personality configurations.

The relationship is consistent with a many-to-one mapping from psychological evaluation to action selection. Enduring traits can alter the relative evaluation of candidate responses without changing which candidate obtains the highest overall score. Factual requirements, normative constraints, and individual preferences jointly shape social decisions, making action convergence particularly plausible in risk management, the fulfillment of responsibilities, and relationship repair. Where several options are acceptable and have similar decision values, personality-related differences can become more visible in implementation parameters: when to request an extension, how to allocate a bonus, how long to attend a social event, or where to establish a relational boundary.

These findings suggest a more informative unit of analysis than whether a response appears characteristic of a particular persona. Trait-specific contributions, composite action scores, and observable action categories can be measured separately and related through scenario-matched comparisons. The target contributions in E02 were generated through the RegistryPMM computation path active during the experiment, demonstrating the traceability of structured personality variables as decision conditions. The resulting research opportunity is to examine **how individual tendencies alter judgment even when they leave the category of the final choice unchanged**.

### 6.3 Experiential Memory and Personality Development: From Retaining History to Changing Future Judgments

Persistent language agents frequently rely on memory retrieval and reflection to maintain behavior across interactions. Generative Agents demonstrated the effectiveness of memory streams, reflection, and planning for this purpose [7]. PMM distinguishes the factual content retained from past experiences from the enduring psychological conditions that shape how subsequent events are interpreted and acted upon. The former describes what happened; the latter specifies how an individual tends to judge new situations after those experiences. These representations interact while following different update rules.

E05 offers a controlled longitudinal comparison of this distinction. Under full learning (C1) and memory-plus-cognitive-state learning with enduring traits frozen (C2), agents with matched initial personalities and experiential environments followed identical trajectories of trust, caution, and mastery. The defining difference was that C1 permitted updates to enduring personality parameters. Relative to C2, C1 showed an additional 0.333-point increase in responsibility and planning across observation waves in the sustained-cooperation environment and an additional 0.250-point increase in the betrayal-and-accountability environment. Both conditions adapted behavior through experience; the additional changes under C1 were concentrated in specific behavioral dimensions associated with enduring-trait updating.

The same experiment revealed continuity alongside plasticity. Sustained cooperation increased mean agreeableness and conscientiousness by 0.100 each, whereas betrayal and accountability decreased agreeableness by 0.100 while increasing conscientiousness by 0.100. Despite these differing trajectories, initial-to-terminal correlations remained 0.9613 and 0.9862 for the two traits, respectively, and all four initial personality configurations retained their high–low group membership. Under betrayal and accountability, full learning also produced two persistent transitions in categorical action strategies, toward stronger rule enforcement and self-protective boundaries.

These observations connect with the TESSERA account of how repeated situational processes contribute to longer-term personality development [6]. Their computational significance lies in the compatibility of individual continuity and gradual change: **continuity does not require personality parameters to remain fixed, and development does not require the loss of prior individual structure**. Personality can change in experience-dependent directions within bounded update rules and subsequently affect behavioral trade-offs. E05 therefore extends the analysis beyond whether an agent remembers an event to whether accumulated experience modifies enduring conditions for future decisions.

### 6.4 Multi-timescale Psychological Dynamics: Stability as Continuity of Update Rules

Personality, cognitive appraisal, and affect operate on different temporal scales. Newly available information may rapidly revise an interpretation of an event; an affective perturbation may alter the immediate state and subsequently subside; enduring traits may change only after evidence accumulates across experiences. PMM organizes these changes through interconnected computational pathways with differentiated update conditions, enabling context-sensitive responses while maintaining continuity of individual psychological states.

The three-stage trajectories in E02 illustrate cognitive adjustment during a single developing interaction. Mean clarification need decreased from 0.3517 to 0.1318 as contextual information accumulated, while the absolute magnitudes of target personality contributions also declined. The personality configuration remained fixed across stages, yet its contribution to the current decision changed. Thus, a stable trait representation can coexist with dynamically changing decision contributions. Psychological stability is expressed through the continuity of the underlying representation and conditional response rules, rather than an invariant contribution magnitude in every situation.

E05 demonstrates a complementary separation over a longer observation window. Short-term perturbations produced substantial changes in affective valence and stress, followed by recovery patterns that were similar across the three PMM-related conditions. In the transient-perturbation environment, enduring A/C parameters remained unchanged. Immediate affective responses and enduring personality characteristics therefore followed different update schedules. Their coexistence makes it possible to preserve long-term individual tendencies while adapting immediate reactions to events and elapsed time.

E03 makes the layered nature of these computations directly observable. Linguistic expression, interpersonal boundary strength, and expression plans varied across assigned affective conditions, whereas effective runtime affective states diverged from their prescribed values. This finding establishes an important measurement requirement: assigned psychological conditions, effective computed states, local decision contributions, and final expressions should be recorded as separate quantities. Their divergences reveal specific locations where a psychological state is transformed or propagated through the architecture, providing concrete targets for calibrating affective dynamics and a more informative assessment than output-text observation alone.

### 6.5 Self-Evaluation Turns Previous Behavior into an Object of Cognition

E04 extends psychological evaluation from external events to records of an agent's own actions. Across 83 complete records, archived scoring classified 30 maintained core choices, 40 implementation adjustments, and 13 core-choice changes; average rubric ratings for discrepancy recognition and source fidelity were 2.916/3 and 2.976/3, versus 2.265/3 for explanation. These summaries describe the full recorded sample. A separate independent reading of ten L08 sequences showed that all seven verified refusal-to-attendance reversals were correctly identified, alongside two compatible conditional choices and one maintained refusal. The convergent evidence supports record-grounded discrepancy recognition while keeping its validation scope identifiable.

The strongest interpretation concerns the distinction between changing an action and changing its underlying guiding considerations. In the verified L08 cases, additional information about a professional opportunity justified time-limited attendance where refusal had originally been expected; the self-evaluation identified the shift while maintaining attention to personal boundaries. A separate L05 example shifted from pursuing an unfulfilled reciprocal commitment toward help after a family hospitalization. The additional audit also identified concrete reconstruction errors involving an afternoon schedule and a private-versus-public communication context. Such cases show that plausible retrospective reasoning and accurate reconstruction of the evidence are analytically distinct, even when an account of changed priorities is coherent.

Computationally, prediction, action, and retrospective evaluation create a second-order monitoring relationship: the system can compare its earlier decision expectations against later recorded responses and identify candidate evidence for future learning. Whole-sample lexical comparisons characterize variation in surface expression, while independently adjudicated decision transitions assess the semantic content of selected self-evaluations. The resulting structure connects monitored behavior with experience-based learning pathways without conflating descriptive ratings, textual divergence, and verified decision recognition. E05 then examines how accumulated experiences, under distinct updating conditions, contribute to subsequent psychological states and behavior.

### 6.6 Positioning among Language-Agent Research: Mechanistic Resolution Rather than the Number of Functions

Recent research has expanded personality and persistent interaction in LLM-based agents from several directions. PersonaLLM investigated the expression of specified personality traits in psychometric assessments and generated text [1]. Serapio-García et al. [3] developed a psychometric framework for evaluating and shaping synthetic personality traits through prompting. Generative Agents used memory, reflection, and planning to sustain social behavior [7], while Cognitive Architectures for Language Agents (CoALA) organized language-agent capabilities in terms of memory, action, and decision-making [20]. Collectively, these studies have advanced the measurement of personality expression, persistent experience management, and modular agent architectures.

PMM contributes to this literature by representing psychological constructs as states that explicitly participate in decision-making and by examining how those states evolve across timescales. Relative to work centered on personality expression, it introduces the intermediate observational levels of trait-specific contributions and composite action scores. Relative to memory-and-reflection architectures, it distinguishes the update conditions of experiential records, transient affect, and enduring personality, and uses matched learning configurations to examine the additional behavioral contribution of trait updating.

The independent psychological module enables these questions to be studied through a common decision interface. Internal event interpretation, personality–situation computation, candidate-action evaluation, and evidence review form the psychological computation process; the General LLM handles user-facing interaction and linguistic realization. The central architectural benefits are **mechanistic observability, persistence of individual states, and comparability across implementations**. E01 also identifies a substantial computational cost: more extensive internal processing required considerably greater cumulative token use despite shorter user-facing responses. Computational depth, quality of expression, and resource requirements therefore remain joint design considerations.

The principal contribution of this study lies in making artificial personality amenable to analysis through the relationships among enduring states, local contributions, integrated decision scores, behavioral outcomes, and experiential updates. The five experiments provide complementary observations at these levels, establishing a research framework that evaluates more than the conversational impression produced by a persona-conditioned model.

### 6.7 Persistent Artificial Psychological Individuals as Testable Computational Systems

The combined findings suggest three connected research dimensions for persistent artificial psychological individuals. The first is **the computational expression of individual differences**: whether enduring personality states produce construct-specific appraisal contributions in matched situations, including cases where final actions converge. The second is **temporal continuity of individual states**: whether psychological responses adjust to changing facts and affect while preserving enduring personality representations and their update rules. The third is **experience-driven development**: whether retained experiences lead to changes in persistent decision tendencies and yield measurable effects on subsequent behavior.

E01 and E02 primarily address the first dimension; E02, E03, and E05 illuminate the second; and E04 and E05 connect records of self-produced behavior with longitudinal experience updating. Viewed together, the experiments articulate computational propositions about how an artificial psychological individual is formed, maintained, and developed, each linked to distinct observable evidence. This cross-experiment interpretation provides a more substantial theoretical contribution than treating personality, emotion, memory, and reflection as a collection of separate capabilities.

The next research steps follow directly from these propositions. Manipulating trait parameters while holding situational inputs constant would trace their influence from local contributions to final actions. Manipulating effective runtime affect under identical external events would test how state changes propagate into decision scores and expression. Varying enduring-trait update rules while holding memory and relationship states fixed would assess behavioral transfer to previously unseen situations. Finally, replacing the General LLM while preserving psychological states would test the continuity of decision structure across language-generation models. Together, these experiments would advance the identification and evaluation of artificial psychological mechanisms.

The broader significance of this research is the investigation of **how an artificial individual can maintain persistent psychological conditions, form judgments under those conditions, and undergo coherent change through experience**. Organizing personality, affect, memory, self-evaluation, and experiential development into interacting yet separately measurable computational processes provides a theoretically grounded and empirically examinable route from persona-conditioned language generation toward persistent artificial psychological individuals.

## 7. Conclusion and Future Directions

### 7.1 Conclusion

This study investigated how artificial intelligence systems can develop persistent individual psychological characteristics and incorporate them into cognition, decision-making, and experiential learning. To address this question, we proposed and implemented PMM, a structured psychological computing framework. PMM organizes long-term personality, dynamic emotion, situational appraisal, self-monitoring of behavior, and experience-dependent updating into interrelated psychological processes with persistent states and explicit computational relationships. Operating as an internal component independent of the General LLM, it contributes to behavioral decision-making through structured psychological conditions. This design enables an artificial individual to maintain its psychological characteristics across interactions, form context-sensitive judgments, and connect accumulated experiences to subsequent behavior.

Methodologically, the study established a computational system encompassing psychological state representation, situation-conditioned appraisal, candidate-action evaluation, behavioral feedback, and long-term state updating. Stable personality provides enduring individual conditions for action appraisal, whereas short-term emotions and cognitive evaluations vary with events and time. Internal computations shape the decision conditions and linguistic expression implemented by the General LLM through structured interfaces. Experience records and self-evaluation further provide evidence for subsequent psychological updates, allowing an individual's interaction history to inform the continuing formation of its behavior.

Five complementary experiments evaluated this system across different temporal scales and levels of observation. The personality-conditioning and cognitive decision-making experiments examined how persistent personality characteristics participate in internal action evaluation and social behavior. The emotion and self-evaluation experiments investigated the computational relationships among transient psychological states, behavioral monitoring, and explanations of action. The longitudinal learning experiment examined experience-dependent changes in stable personality states, continuity of individual differences, and subsequent behavioral adjustment. Together, these experiments established a multilevel evaluation approach spanning internal psychological states, dynamic cognitive computations, observable social behavior, and long-term individual development.

Three overarching findings emerge from this evidence. First, stable individual characteristics can influence internal appraisal and action trade-offs, with measurable relationships between psychological contributions and final behavior. Second, the persistence and adaptability of psychological states can be jointly supported by maintenance and updating mechanisms operating at different temporal scales, allowing an individual to respond to changing circumstances while preserving continuity in its psychological and behavioral characteristics. Third, accumulated experience can contribute to the development of enduring decision conditions, providing a computational basis for personality-related change over extended interaction.

Accordingly, this research establishes artificial personality as a subject that can be investigated through **states, computational processes, behavior, and developmental trajectories**, and provides an architectural implementation and empirical approach for doing so. More broadly, it outlines a research pathway for modeling persistent artificial individuals: an individual maintains relatively stable psychological characteristics, applies them when interpreting situations and making decisions, interacts with its environment through behavior, and gradually modifies its psychological states and future behavioral tendencies through accumulated experience.

## 7.2 Future Research Directions

The framework developed in this study provides a foundation for investigating the continued learning and development of artificial individuals. Future work will focus on the acquisition of psychological characteristics through experience, the organization and transformation of experiential memory, the formation of intrinsic motivation, and the continuity of individual identity across language models and real-world environments. Together, these directions aim to extend psychological computing from persistent-state decision support toward long-term interactive agents that learn from their own experiences, develop behavioral tendencies, and initiate goal-directed activity.

### 7.2.1 From Structured Personality Representations to Learnable Psychological Mechanisms

How enduring personality characteristics emerge from experience is a central question for subsequent research. The present framework provides explicit personality-state representations, updating rules, and interfaces through which personality contributes to decision-making. These components establish a basis for investigating mechanisms through which stable psychological characteristics can be learned rather than specified exclusively in advance.

Future research will combine longitudinal interaction records, behavioral outcomes, situational information, and evaluative feedback to learn stable personality representations, context-dependent judgment patterns, and behavioral preferences. Potential computational developments include trainable psychological-state encoders, personality-estimation networks, and dynamic decision-weight generators. Long-term personality learning and short-term emotional and cognitive updating will be coordinated within a multiscale temporal framework.

Changes in enduring traits will be informed by evidence accumulated across situations, while immediate behavior will remain sensitive to the current state and context.

A further priority is to determine how learned personality representations affect the computations underlying actual decisions. By holding situations constant while varying psychological states and comparing candidate-action evaluations, researchers can identify the influence of individual characteristics on judgment and action trade-offs. Tests in previously unseen situations and extended interaction sequences can then assess the stability, transferability, and developmental properties of the learned representations.

This direction will progressively extend PMM from explicit parameter-based conditioning toward neural psychological models capable of learning from experience, making the formation, maintenance, and application of psychological characteristics jointly trainable and empirically testable.

### 7.2.2 Memory Formation, Experience Integration, and the Development of Behavioral Habits

The long-term development of an artificial individual depends on the continuing interpretation and organization of experience. Future work will develop a multilayer memory system encompassing episodic records, generalized knowledge, relational experiences, and habitual behavior, while investigating how these forms of memory emerge, are updated, and interact over extended periods.

An initial challenge is to identify, segment, and encode continuous events. Memory representations should preserve event facts, contextual conditions, individual appraisals, actions taken, and their outcomes so that relevant experiences can be retrieved during subsequent judgments. Evidence linking, contradiction detection, and memory consolidation can then support the gradual transformation of recurring experiences into more stable knowledge, relational understanding, and behavioral tendencies.

A second priority is to establish clearer connections between memory use and specific psychological functions. Emotionally salient experiences may inform later event appraisal and emotion regulation; experiences associated with enduring behavioral preferences may contribute to personality learning; and information about goals and action outcomes may support motivational processes and planning. The system should also maintain distinct representations of its own experiences, its understanding of other individuals, and the state of relationships between them, allowing information from different sources to be retrieved and applied appropriately.

The formation of behavioral habits represents an important extension of this work. By examining repeated action choices and outcomes in recurring contexts, future studies can investigate how relatively stable context-response tendencies emerge and how they interact with personality, momentary emotions, and deliberative planning. Manipulations of contextual cues, feedback, and accumulated experience can be used to evaluate habit acquisition, persistence, and modification.

This line of research will allow memory to contribute to the development of enduring psychological states and behavioral patterns, extending its role beyond the storage and retrieval of historical information and providing a computational basis for experience-dependent individual development.

### 7.2.3 Autonomous Motivation and Goal Formation

Building on persistent psychological states and experiential learning, autonomous motivation constitutes a major direction for future research. An artificial individual may need to do more than respond to external inputs: it may also form intentions to act based on its internal state, changes in the environment, and accumulated experience. Understanding how internally evaluated conditions give rise to goals and continue to influence behavior presents a new computational research problem.

Future studies will examine relationships among internal needs, current psychological states, existing goals, environmental opportunities, and anticipated outcomes during motivation formation. Personality may shape preferences among potential goals, emotional and cognitive appraisal may alter their immediate salience, and memory may supply evidence about previous actions and consequences. An independent motivational mechanism could then generate, evaluate, and select candidate goals before passing selected intentions to an execution system.

These processes also require models of how motivation changes over both real time and interaction time. Some goals may gradually gain priority, whereas others may be revised in response to new information, unsuccessful actions, or competing needs. Organizing internal state updates, goal appraisal, action execution, and outcome feedback into a continuous process will make it possible to study how motivations arise, persist, compete, and change.

Experimentally, agents could be observed in environments without immediate external task instructions to determine whether they generate contextually meaningful goals. Evaluation would address the appropriateness and persistence of goal selection, the ability to revise goals, and consistency with the individual's psychological characteristics. Research should also examine how self-generated goals are coordinated with external tasks, permissions, and resource constraints so that proactive behavior remains interpretable, controllable, and compatible with continued learning.

This direction will connect persistent psychological conditions more directly to autonomous action, extending the study of artificial individuals from responses to external events toward internally initiated behavior.

#### 7.2.4 Cross-Model Psychological Continuity and Neural Representations of Individuality

Another research opportunity offered by an independent psychological computing framework is the preservation and application of persistent individual states across different General LLMs. As foundation models continue to change in capability and implementation, maintaining the psychological characteristics of the same artificial individual across model transitions will become an important issue for persistent agents.

Future studies will compare how different language models behave when supplied with identical psychological states and decision conditions, examining the effects of model replacement on situational judgment, action selection, personality expression, and language generation. Building on these comparisons, further research will investigate relationships between the psychological computing module and internal language-model representations, including trainable lightweight adapters, hidden-state activation steering, and mappings of personality representations across models.

A central requirement is to establish measurable criteria for individual continuity. After replacing the General LLM, does an artificial individual retain similar value judgments, behavioral preferences, and social decision-making tendencies? Can its psychological state continue to develop along a coherent trajectory following new experiences? Does language generated by different models faithfully implement the same internally selected action? These questions provide concrete evaluation targets for cross-model continuity.

Such work will clarify the relationship between continuity in an artificial individual's psychological state and the computational realization of its underlying language model, while extending the applicability of persistent personality representations and learned experience across model architectures.

#### 7.2.5 Long-Term Naturalistic Interaction and Validation of Psychological Computing Systems

Future research will progressively extend controlled social scenarios to long-term naturalistic interaction environments, integrating real-world time, continuous events, multimodal information, and autonomous action into a more comprehensive experimental setting for artificial individual development.

Longitudinal interaction offers a wider range of conditions under which psychological change can be studied, including recurring relationships, sustained tasks, interpersonal conflicts, environmental changes, and repeated behavioral feedback. These settings will enable further investigation of how personality, emotion, memory, habits, and motivation jointly shape behavior and how enduring individual characteristics are maintained or modified through experience.

Multimodal perception and agent tools will also expand the range of inputs and actions available to the psychological computing system. Future implementations may transform text, images, video, and environmental events into structured contextual evidence, while obtaining new experiences through tool use, information gathering, and interaction with external environments. The central research question will concern how psychological computations contribute to a continuous perception-appraisal-action-feedback process and how the resulting experiences subsequently affect the individual's state.

Evaluation will combine longitudinal behavioral measures, assessments of psychological constructs, independent human feedback, and internal computational traces. These sources can support a multilevel assessment of individual stability, behavioral adaptability, learning effectiveness, and psychological change. To address the computational demands of sustained operation, future work will also investigate adaptive scheduling of internal reasoning, memory retrieval, and decision auditing, allowing more intensive psychological computation to be used when warranted by the situation.

Progressive expansion from controlled experiments to naturalistic environments will provide stronger tests of psychological computing systems under complex, open-ended, and extended interaction conditions, while informing the continued refinement of their computational mechanisms.

### 7.3 Long-Term Research Outlook

This study establishes an initial computational and empirical foundation for persistent artificial psychological systems. Future research will build on this foundation through personality learning, experiential memory, self-evaluation, behavioral habit formation, and autonomous motivation. Cross-model validation and long-term interactive studies will then support the integration of these mechanisms into artificial intelligence systems with enduring individual characteristics.

The central question guiding this research agenda is: **How can an artificial individual preserve its characteristic dispositions over extended experience while continuing to learn, revise its judgments, develop new behavioral tendencies, and progressively acquire the capacity to initiate and organize its own actions?**

Addressing this question requires the joint study of continuity and development. Enduring psychological states provide a relatively stable basis for judgment; experiential learning allows that basis to be revised in light of accumulated history; memory and self-evaluation connect prior events to current appraisal; and motivational mechanisms link persistent psychological conditions to future action. As these computational processes become more fully developed, artificial individuals may exhibit behavioral trajectories characterized by longer-term historical dependence, adaptation, and individual differentiation.

Accordingly, future work will use persistent psychological states and experiential learning as a foundation for investigating artificial psychological systems capable of autonomous goal formation, long-term behavioral continuity, and adaptation across environments. This agenda offers a path toward more systematic computational models and experimental methods for individualized artificial intelligence, long-term social agents, and artificial psychological mechanisms.